%% file: acl_latex.tex
\documentclass[11pt]{article}

\usepackage{acl}
\usepackage[T1]{fontenc}
\usepackage[utf8]{inputenc}
\usepackage{times}
\usepackage{latexsym}
\usepackage{inconsolata}
\usepackage{microtype}
\usepackage{amsfonts}
\usepackage{amsmath,amssymb}
\usepackage{xcolor}
\usepackage{graphicx}
\usepackage{booktabs}
\usepackage{multirow}
\usepackage{colortbl}
\usepackage{enumitem}
\usepackage[most]{tcolorbox}
\tcbuselibrary{breakable,skins,theorems}
\usepackage{xspace}
\usepackage{pifont}

\definecolor{dartgreen}{HTML}{00693e}
\definecolor{caseblue}{HTML}{1F5A94}
\definecolor{casegreen}{HTML}{287A55}
\definecolor{casepurple}{HTML}{6C4A8E}
\definecolor{caseorange}{HTML}{A65A1F}
\definecolor{caseteal}{HTML}{13747D}
\definecolor{casered}{HTML}{9B3446}
\colorlet{casebody}{black}

\tcbset{
  utilmemexample/.style={
    enhanced,
    breakable,
    boxrule=0.55pt,
    arc=1.5mm,
    left=5pt,
    right=5pt,
    top=4pt,
    bottom=4pt,
    before skip=6pt,
    after skip=7pt,
    fonttitle=\bfseries\small,
    fontupper=\small,
    coltitle=white,
    coltext=casebody,
    colback=blue!2!white
  },
  utilmemdistractor/.style={
    utilmemexample,
    colframe=caseorange,
    colback=caseorange!4!white,
    colbacktitle=caseorange
  },
  utilmemfailure/.style={
    utilmemexample,
    colframe=casered,
    colback=casered!3!white,
    colbacktitle=casered
  }
}

\newcommand{\cmark}{\ding{51}}
\newcommand{\xmark}{\ding{55}}
\newcommand{\pmark}{$\bullet$}
\newcommand{\ourmethod}{\textsc{UtilMem}\xspace}

\title{\ourmethod: Benchmarking Evidence Utilization in Long-Term Conversational Memory}

\author{
  Peijun Qing$^{1}$ \qquad Fobo Shi$^{2}$ \qquad Soroush Vosoughi$^{1}$ \\
  $^{1}$Dartmouth College \qquad
  $^{2}$Wuhan University \\
  \texttt{peijun.qing.gr@dartmouth.edu}
}

\begin{document}
\maketitle

\input{sections/abstract}

\input{sections/introduction}

\input{sections/related_work}

\input{sections/benchmark}

\input{sections/experiments}

\input{sections/Results}

\input{sections/conclusion}

\input{sections/Limitations}

\section*{Ethics Statement}

\textsc{UtilMem} is intended for research evaluation and is not suitable for clinical diagnosis, treatment, crisis triage, or other clinical decision-making. Its Mental Wellness domain is constructed from an existing, publicly accessible Reddit-derived dataset that contains first-person discussions of mental-health experiences. Public accessibility should not be interpreted as consent for unrestricted reuse. We use these posts as inputs to an LLM-mediated transformation pipeline and do not treat self-reported conditions as clinically verified diagnoses. Although source-platform usernames and post identifiers are not intentionally included as dedicated metadata, transformed conversations may retain contextual details; we therefore do not claim that re-identification is impossible. Users should not attempt to identify or contact the original authors.

\section*{Acknowledgment}

This research was supported in part by the National Science Foundation under Grant No. 2452367.

\bibliography{custom}

\clearpage
\appendix

\input{sections/appendix_data_stats}
\input{sections/appendix_imple_detail}

\input{sections/appendix_data_preprocess}

\input{sections/appendix_domain_examples}

\input{sections/appendix_judge_oracle}

\input{sections/appendix_data_syn_prompt}

\end{document}

%% file: sections/abstract.tex
\begin{abstract}

Long-term memory is increasingly important for conversational agents, yet existing benchmarks primarily measure memory through pointwise factual recall: whether a system can recover isolated facts or event-level details from prior interactions. Real-world memory use, however, often requires a more demanding capability: integrating distributed, implicit, and noisy evidence across extended interaction histories into coherent, task-oriented outputs. We call this capability \emph{memory utilization}. Here, we introduce \textsc{UtilMem}, a diagnostic benchmark comprising 1,717 instances across five domains, designed to evaluate four underexplored aspects of memory utilization: reasoning over dense histories, identifying implicitly relevant memories, synthesizing distributed evidence into summaries, analyses, or plans, and resisting interference from semantically similar distractors. Evaluating a diverse set of retrieval-based and memory-augmented systems, we find that strong performance on conventional factual-memory benchmarks does not reliably translate into effective memory utilization. Moreover, retrieval alone is insufficient: even when relevant evidence is successfully recovered, systems frequently fail to integrate information across sessions or to distinguish useful evidence from plausible distractors. These findings expose a substantial gap between accessing stored information and using it effectively, and suggest that progress in long-term conversational memory will require architectures that explicitly support evidence integration and robustness to retrieval interference. Code is available at \href{https://github.com/peijunallin/UtilMem}{https://github.com/peijunallin/UtilMem}.

\end{abstract}

%% file: sections/introduction.tex
\section{Introduction}

Large language model (LLM) assistants are increasingly deployed in long-horizon, multi-session settings such as personal companions, tutors, financial advisors, and health coaches, where users expect the system to remember past interactions and act on them~\citep{zhong2024memorybank,packer2023memgpt,chhikara2025mem0,qing2026cluster}. This regime has motivated a wave of memory-augmented architectures that compress, index, and retrieve from chat histories~\citep{packer2023memgpt,xu2025amem,gutierrez2024hipporag,wang2025mirix,diao2025temporal,qing-etal-2026-tailoring}, alongside benchmarks that aim to measure how well these systems support long-term interaction~\citep{maharana2024locomo,wu2024longmemeval,hu2026memoryagentbench,tan2025membench,bian2026realmem}. Yet what existing benchmarks measure is substantially narrower than what deployed memory systems are asked to do.

\begin{figure}[t]
    \centering
    \includegraphics[scale=0.4]{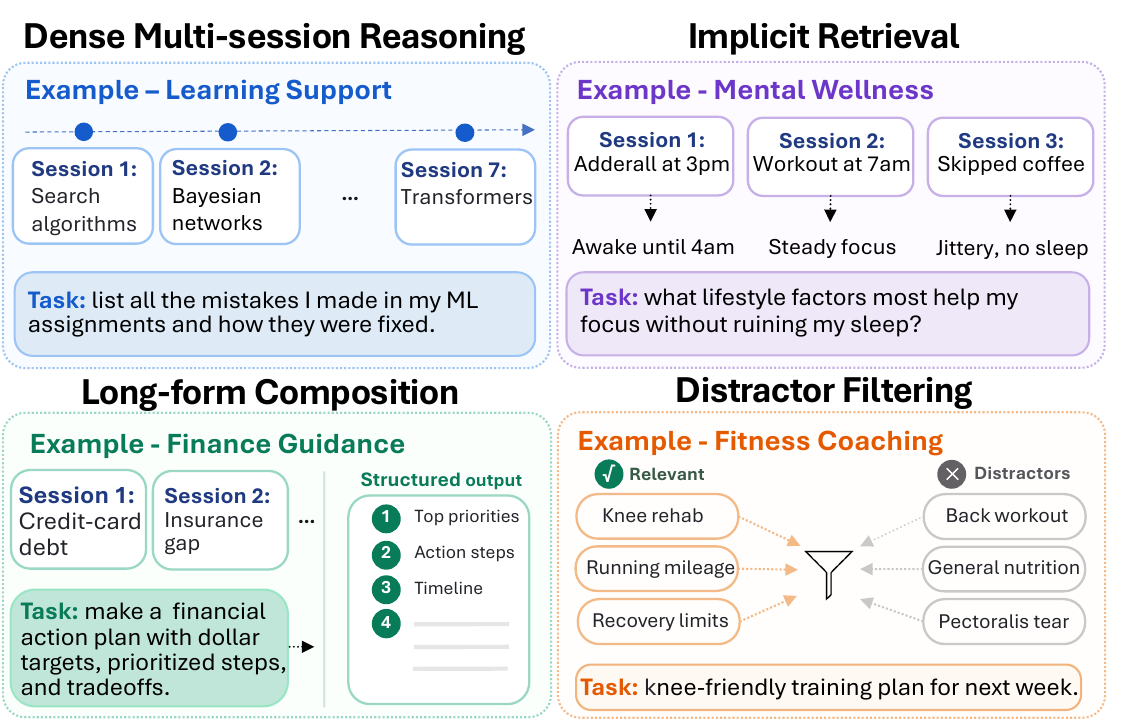}
    \caption{Memory utilization tasks in \textsc{UtilMem}.}
    \label{fig:intro}
    \vspace{-6mm}
\end{figure}

Existing long-term memory benchmarks are primarily evaluated through pointwise factual recall against atomic gold answers. Benchmarks such as LoCoMo~\citep{maharana2024locomo} and LongMemEval~\citep{wu2024longmemeval} primarily evaluate whether a system can retrieve a relevant span from conversation history and answer a narrowly scoped question. However, real memory usage is rarely a retrieval-only problem. Current evaluations leave several critical aspects underconstrained: \textbf{(1) localized evidence}, where supporting information is concentrated within a small number of sessions and long-horizon reasoning collapses into local retrieval; \textbf{(2) limited distractor pressure}, where distractors are weak or topically distant, allowing systems to improve performance simply by increasing top-$k$ retrieval; \textbf{(3) short-form outputs}, where tasks require only atomic factual responses rather than coherent long-form synthesis; and \textbf{(4) shortcut-friendly evaluation}, where strong benchmark performance can be achieved through event-level retrieval, atomic fact extraction, or aggressive memory compression that converts raw interaction histories into compact summaries or structured slots. As a result, these limitations undermeasure an essential goal of long-term memory that we operationalize as \emph{memory utilization}: the ability to synthesize distributed, implicit, and noisy evidence from long interaction histories into coherent task-oriented responses.

To systematically evaluate this requirement, we introduce \textsc{UtilMem}, which targets four challenges (see Figure~\ref{fig:intro}) largely absent from recall-oriented evaluations: \textbf{Dense Multi-session Reasoning}, where evidence is distributed across temporally separated sessions and topics; \textbf{Implicit Retrieval}, where systems recover latent user preferences, behavioral patterns, and evolving states rather than explicit facts; \textbf{Long-form Composition}, where fragmented and partially redundant evidence must be organized into coherent outputs; and \textbf{Distractor Filtering}, where semantically similar but functionally irrelevant sessions create retrieval and generation interference.

\textsc{UtilMem} is built from realistic multi-session interaction trajectories spanning learning support, finance guidance, mental wellness, fitness coaching, and document analysis. We generate evidence-dependent tasks requiring cross-session integration and apply counterfactual filtering to retain only instances whose answer quality degrades along the sampled sequential evidence-removal path. We further inject adversarial distractors that overlap lexically and semantically with the evidence while differing in intent, forcing systems to separate grounded memory from plausible interference.

We evaluate retrieval-augmented and memory-augmented systems on \textsc{UtilMem}. Results reveal a consistent gap between memory access and memory utilization. Systems performing strongly on factual memory benchmarks degrade substantially when tasks require cross-session reasoning under retrieval noise. Even when relevant evidence is retrieved, models often fail to compose coherent grounded outputs or instead rely on semantically plausible distractors that produce confident but unsupported generations. For the evaluated systems and configurations, these findings diagnose failure modes that recall-oriented evaluation does not expose; they do not establish a universal mechanism for all long-term memory systems. Our contributions are summarized as follows:
\begin{itemize}
\item A benchmark for evaluating memory utilization in long-term conversational agents.

\item A scalable synthesis pipeline that produces evidence-sensitive evaluation instances with adversarial retrieval interference.

\item An empirical evaluation showing that strong performance on existing factual memory benchmarks does not guarantee robust memory utilization.
\end{itemize}

%% file: sections/related_work.tex
\begin{table*}[t]
\centering
\renewcommand{\arraystretch}{1.15}
\resizebox{2\columnwidth}{!}{%
\begin{tabular}{lccccccc}
\toprule
\multirow{2}{*}{\textbf{Benchmark}} & \multirow{2}{*}{\textbf{\#Q}} & \multirow{2}{*}{\textbf{Context Depth}} & \multirow{2}{*}{\textbf{Eval. Metric}} & \multicolumn{4}{c}{\textbf{Memory Utilization Dimensions}} \\
\cmidrule(lr){5-8}
 & & & & \textbf{D1: Dense} & \textbf{D2: Implicit} & \textbf{D3: Long-form} & \textbf{D4: Distractor} \\
 & & & & \textbf{Multi-session} & \textbf{Retrieval} & \textbf{Composition} & \textbf{Filtering} \\
\midrule
LoCoMo~\citep{maharana2024locomo}
  & 7{,}512
  & $\sim$9k
  & F1 / BLEU / ROUGE
  & \cmark & \xmark & \pmark & \xmark \\
LongMemEval~\citep{wu2024longmemeval}
  & 500
  & 115k--1.5M
  & QA Acc. / Recall@k / NDCG@k
  & \pmark & \xmark & \xmark & \xmark \\
MemoryBank~\citep{zhong2024memorybank}
  & 194
  & $\sim$5k
  & Manual + LLM judge
  & \xmark & \xmark & \xmark & \xmark \\
MemBench~\citep{tan2025membench}
  & 1{,}039
  & 1k--100k
  & MCQ Acc. / Recall@10
  & \cmark & \cmark & \xmark & \pmark \\
MemoryAgentBench~\citep{hu2026memoryagentbench}
  & 2{,}071
  & 103k--1.44M
  & Accuracy / Recall@5 / F1
  & \cmark & \pmark & \cmark & \xmark \\
RealMem~\citep{bian2026realmem}
  & 1{,}415
  & $\sim$269k
  & QA Score / Recall@k / NDCG@k
  & \cmark & \cmark & \xmark & \xmark \\
ConvoMem~\citep{pakhomov2025convomem}
  & 75{,}336
  & 1k--3M
  & LLM-judge Accuracy
  & \pmark & \cmark & \xmark & \xmark \\
\midrule
\textbf{\textsc{UtilMem} (Ours)} & \textbf{1{,}717} & \textbf{80k--220k} & \textbf{Rubric LLM-judge} & \cmark & \cmark & \cmark & \cmark \\
\bottomrule
\end{tabular}}
\caption{Comparison of long-term conversational memory benchmarks. \ding{51}: fully supported; $\bullet$: partially supported; \ding{55}: not supported. Under our operationalization, \textsc{UtilMem} is the only benchmark in this comparison that jointly targets all four utilization dimensions.}
\label{tab:benchmark_comparison}
\end{table*}

\section{Related Work}

\subsection{Benchmarks for Long-Term Conversational Memory}

Long-term memory benchmarks have largely evolved within a single evaluation paradigm: factual question answering over multi-session dialogue histories. Early work established the format with multi-day chat probes \citep{du2024perltqa, kim2024dialsim, zhong2024memorybank} and human--human conversations spanning single-hop, multi-hop, temporal, and adversarial questions \citep{maharana2024locomo}. LongMemEval \citep{wu2024longmemeval} consolidated this protocol around five recall-oriented abilities for user--assistant interactions, namely information extraction, multi-session reasoning, temporal reasoning, knowledge updates, and abstention. Recent extensions diversify the protocol along several axes: broader question taxonomies that include reflective memory and observer scenarios \citep{tan2025membench}, operation-level decomposition that localizes failures within the memory pipeline \citep{chen2025halumem}, finer cognitive competencies under incremental multi-turn formats \citep{hu2026memoryagentbench}, project-oriented interactions with interwoven queries \citep{bian2026realmem}, and large-scale conversational corpora with implicit cues \citep{pakhomov2025convomem}.

Despite this breadth, most benchmarks emphasize pointwise factual recall or task-specific short-answer signals. Some include multi-session reasoning or summarization, but none jointly couples dense, implicit evidence with long-form task-oriented composition and semantically similar distractor filtering. As summarized in Table~\ref{tab:benchmark_comparison}, none of the benchmarks compared jointly covers the four utilization dimensions we identify: dense multi-session evidence, implicit retrieval, long-form composition, and distractor filtering. \textsc{UtilMem} targets this gap by evaluating memory utilization: long-form composition over distributed, implicit evidence under semantically similar distractors. Noise sensitivity is also studied in retrieval-augmented generation. Prior work shows that irrelevant context can distract language models~\citep{shi2023distract}, that useful evidence can be underused depending on its position in a long context~\citep{liu2024lostinthemiddle}, and that RAG systems vary in noise robustness, negative rejection, information integration, and counterfactual robustness~\citep{chen2024rgb}. \textsc{UtilMem} extends this diagnostic setting to long-term conversational memory: its distractors are complete conversation sessions that deliberately share lexical and semantic cues with the evidence while remaining functionally irrelevant to the user's request, and the output requires synthesis across multiple sessions rather than passage-level extraction.

\subsection{Memory Systems for Long-Term Interaction}

A parallel line of work develops explicit memory systems that compress, organize, and retrieve from interaction histories \citep{wang2024lifespan, zhang2024memorysurvey, du2025rethinking, liu2025foundationagents, mei2025context}. Existing designs cluster into three directions. The first stores experiences as linear or hierarchical streams with policies for retention and forgetting \citep{packer2023memgpt, zhong2024memorybank, liang2023scm, park2023generative}. The second imposes structure through graphs, trees, or temporal indices to support compositional retrieval and update \citep{xu2025amem, chhikara2025mem0, rasmussen2025zep, rezazadeh2025tree}. The third integrates heterogeneous memory types, including parametric, activation, procedural, and multi-agent memory, under unified lifecycle control \citep{li2025memos, kang2025memory}. Retrieval-augmented generation and its graph-augmented variants remain widely used as memory backends \citep{lewis2020rag, edge2024graphrag, gutierrez2024hipporag, guo2024lightrag}.

These systems target distinct facets of memory, including storage efficiency, update consistency, relational structure, and retrieval precision. Their reported gains, however, are often evaluated with recall-oriented QA signals, leaving open whether architectural progress translates to tasks that require evidence integration rather than fact lookup. \textsc{UtilMem} provides such a testbed.

%% file: sections/benchmark.tex
\begin{figure*}[t]
    \centering
    \includegraphics[width=\textwidth]{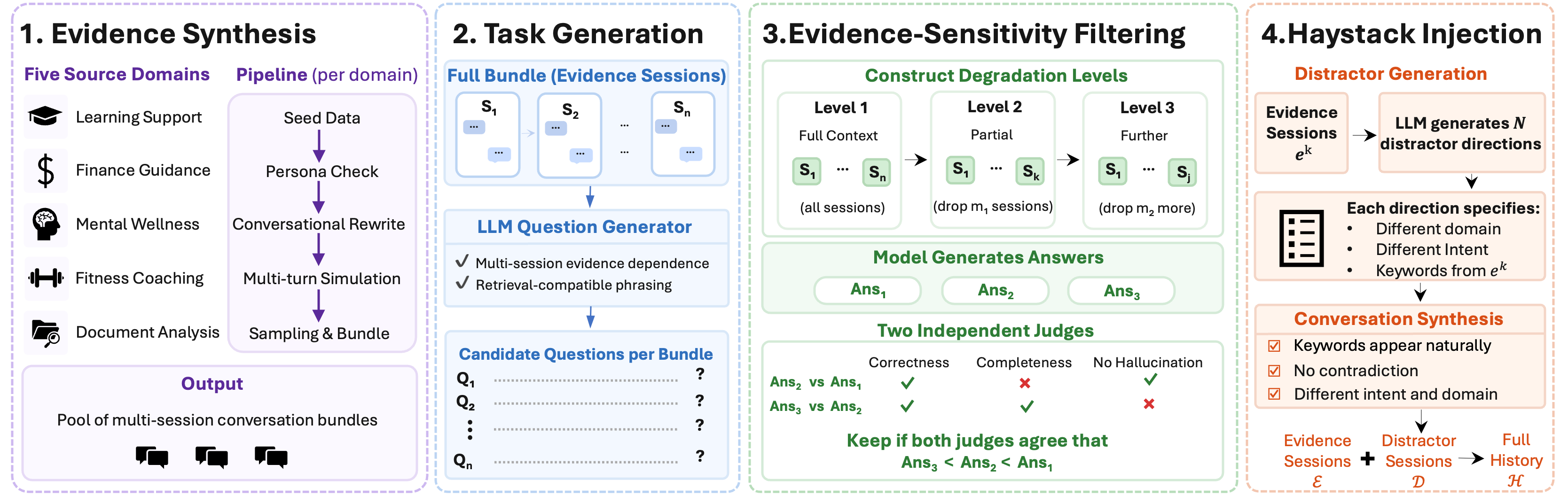}
    \caption{Overview of the \textsc{UtilMem} construction pipeline, including multi-session evidence synthesis, task generation, evidence-sensitivity filtering via counterfactual degradation, and adversarial distractor injection for haystack construction.}
    \label{fig:data_pipe}
\end{figure*}

\section{Benchmark Construction}
\label{sec:construction}

This section starts with the formulation of our task. As illustrated in Figure~\ref{fig:data_pipe}, we first synthesize realistic multi-session interaction histories (\S\ref{sec:evidence-construction}), then generate evidence-dependent questions (\S\ref{sec:task_generation}), filter for evidence sensitivity (\S\ref{sec:evidence_filtering}), and finally inject adversarial distractors to construct retrieval-interfering haystacks (\S\ref{sec:haystack_injection}).
\paragraph{Problem Formulation}
\label{sec:formulation}
 
A benchmark instance is a tuple $(\mathcal{H}, q, \mathcal{E})$, defined as follows. Let $\mathcal{H} = \{s_1, s_2, \ldots, s_{|\mathcal{H}|}\}$ denote a user's complete interaction history, consisting of $|\mathcal{H}|$ chat sessions.
Each session $s_i$ is a multi-turn dialogue between a user and an assistant:
\begin{equation}
    s_i = \bigl[(u_i^{(1)}, a_i^{(1)}),\; (u_i^{(2)}, a_i^{(2)}),\; \ldots,\; (u_i^{(T_i)}, a_i^{(T_i)})\bigr],
\end{equation}
where $(u_i^{(t)}, a_i^{(t)})$ is the $t$-th turn consisting of a user utterance and an assistant response, and $T_i$ is the number of turns in session $s_i$. The history $\mathcal{H}$ is partitioned into two disjoint subsets:
\begin{equation}
    \mathcal{H} = \mathcal{E} \cup \mathcal{D}, \quad \mathcal{E} \cap \mathcal{D} = \emptyset,
\end{equation}
where $\mathcal{E} = \{e_1, \ldots, e_K\}$ is the set of \emph{evidence sessions} containing information relevant to the evaluation task, and $\mathcal{D} = \{d_1, \ldots, d_L\}$ is the set of haystack sessions. Associated with each instance is a \emph{question pool}
\begin{equation}
    \mathcal{Q} = \{q_1, q_2, \ldots, q_n\},
\end{equation}
where each $q_j \in \mathcal{Q}$ is an open-ended question designed to require synthesis across multiple evidence sessions in $\mathcal{E}$. At evaluation time, for each question $q \in \mathcal{Q}$ a system receives the history $\mathcal{H}$ and the question $q$, and produces a response $\hat{a}$.

\subsection{Evidence Synthesis}
\label{sec:evidence-construction}

To ensure realism and diversity, we seed \ourmethod with content drawn from five public datasets, each capturing a distinct mode of user--assistant interaction. Learning Support draws from StudyChat~\citep{mcnichols2025studychat}, a corpus of student conversations with an AI tutor across seven course topics in an undergraduate AI class. Finance Guidance draws from Personal-Finance-Queries~\citep{personalfinance_redditqa}, a Reddit-sourced question--answer dataset covering budgeting, investing, debt, insurance, and savings.
Mental Wellness draws from a Reddit mental health corpus~\citep{mentalhealth} containing first-person posts from subreddits such as \textit{depression}, \textit{ADHD}, \textit{OCD}, and \textit{ptsd}.
Fitness Coaching draws from a fitness question--answer dataset\footnote{\url{https://huggingface.co/datasets/its-myrto/fitness-question-answers}} covering exercise programming, nutrition, and recovery. Document Analysis draws from S\&P~500 10-K filings\footnote{\url{https://huggingface.co/datasets/jlohding/sp500-edgar-10k}}, where each filing is decomposed into standardized items (e.g., Business, Risk Factors, Legal Proceedings). Each domain is processed through a synthesis pipeline that shares a common four-step structure.
(1) Sampling. Source items are grouped into bundles $B = \{s_1, \ldots, s_K\}$ representing a single user's history, with sessions drawn from different topical subareas within the domain to encourage breadth of evidence. (2) Persona consistency. An LLM judge rejects bundles containing contradictions in stable user attributes such as conflicting ages or mutually exclusive circumstances, ensuring each bundle plausibly originates from one individual.
(3) Conversational rewriting. Source items that are not already in dialogue form are rewritten into natural user--assistant exchanges, after which a back-and-forth loop between an agent model and a user model expands each seed into a multi-turn session while preserving all concrete details. (4) Deduplication. Bundle-level deduplication ensures no two samples share the same source-item set. The output is a pool of conversation bundles, each a plausible multi-session interaction history for a single user. Detailed per-domain processing is provided in Appendix~\ref{sec:appendix-domain-synthesis}.

\subsection{Question Generation}
\label{sec:task_generation}

For each bundle $B$, we generate candidate questions satisfying two properties.
\textit{Multi-session dependence} is a generation target: a candidate question $q$ should benefit from aggregating information distributed across the sessions in $B$. \textit{Retrieval-compatible phrasing} requires that $q$ uses terms appearing in the conversation text (e.g., \textit{my AI course assignments}, \textit{credit card debt}), ensuring that retrieval-based systems can operate on the questions as posed. Beyond these two properties, we additionally require questions to be \textit{deterministic and non-judgmental}, asking for organization, integration, listing, or planning operations whose outputs are grounded directly in the conversation content, rather than for subjective diagnoses (e.g., \textit{weaknesses}, \textit{most important takeaway}) that would yield unstable answers across model runs. A language model receives the complete text of all sessions in $B$ with internal session labels stripped, and emits ten candidate questions per bundle.
Domain-specific instantiations of the prompt guide the question types appropriate to each domain.
The full system prompt and the per-domain templates are provided in Appendix~\ref{sec:appendix-qgen-prompts}.

\subsection{Evidence-Sensitivity Filtering}
\label{sec:evidence_filtering}

Not all candidate questions genuinely require all sessions. Some are answerable at comparable quality from a subset due to information redundancy or underspecification.
We filter these via a counterfactual degradation test. For each bundle with $K$ evidence sessions, we construct three context levels by progressively removing sessions:
\begin{equation}
    \mathcal{E}^{(1)} = \mathcal{E}, \quad
    \mathcal{E}^{(2)} = \mathcal{E}^{(1)} \setminus R_1, \quad
    \mathcal{E}^{(3)} = \mathcal{E}^{(2)} \setminus R_2,
\end{equation}
where $R_1, R_2 \subset \mathcal{E}$ are randomly selected removal sets.
The removal sizes $|R_1|$ and $|R_2|$ are adapted proportionally to $K$, targeting approximately 50\% total removal while ensuring $|\mathcal{E}^{(3)}| \geq 1$.
A target model generates answers $\hat{a}^{(1)}, \hat{a}^{(2)}, \hat{a}^{(3)}$ at each context level.
Two independent judge models then perform pairwise comparisons, each rendering a binary verdict on whether the degraded answer is worse than the fuller-context answer.
We verify \textit{monotonic degradation} and retain a question only if both judges agree that
\begin{equation}
    \text{quality}(\hat{a}^{(1)}) > \text{quality}(\hat{a}^{(2)}) > \text{quality}(\hat{a}^{(3)}),
\end{equation}
which requires four concordant binary verdicts from two judges across two adjacent comparisons.
This conservative sampled degradation test retains questions whose answer quality decreases along one sequential removal path, $R_1$ followed by $R_2$. It establishes evidence sensitivity under the sampled removals. The system prompt used for the pairwise judges is provided in Appendix~\ref{sec:appendix-judge-prompt}. Human verification and case study are provided in Appendix~\ref{app:oracle-justification}.

\subsection{Haystack Construction}
\label{sec:haystack_injection}

Weak distractors allow systems to succeed through shallow retrieval, reducing evaluation to recall rather than utilization. To stress-test robustness under retrieval interference, we construct adversarial distractors that remain semantically close to the evidence sessions $\mathcal{E}$ while being functionally irrelevant to the target question.

\paragraph{Distractor synthesis.}
For each evidence session $e_k \in \mathcal{E}$, we generate distractors in two stages. First, an LLM reads $e_k$ and produces $N$ distractor directions sampled from curated cross-domain families. Each direction specifies a target domain, a conversational intent, a domain distinguisher, and 5--10 keywords shared with $e_k$. The shared keywords preserve lexical overlap with the evidence, while the new domain and intent shift the underlying semantics. Second, each direction is instantiated into $M$ multi-turn conversations with turn counts sampled uniformly from $[\tau_{\min}, \tau_{\max}]$. The synthesis prompt enforces three constraints: (1) shared keywords must appear naturally, (2) domain-specific terminology must dominate the conversation so the distractor intent is immediately recognizable, and (3) no evidence content may be copied or contradicted. As a result, distractors remain retrieval-plausible but answer-irrelevant.

\paragraph{Haystack assembly.}
Each evidence session generates $N \cdot M$ in-domain distractors, and all distractors are aggregated into a single haystack $\mathcal{D}$. To additionally simulate the heterogeneous content found in real user histories, the haystacks of the five domains are then merged so that each sample's history $\mathcal{H} = \mathcal{E} \cup \mathcal{D}$ contains its own in-domain evidence and strong distractors alongside cross-domain sessions that act as weak distractors. Real timestamps from the fitness domain are used as temporal anchors, while sessions from other domains receive synthesized timestamps that preserve their relative ordering. Distractors are distributed throughout the timeline before the full history is sorted chronologically for evaluation.

This construction separates evidence dependence from retrieval robustness. The evidence-sensitivity filter selects questions that degrade along the sampled removal path, while the adversarial haystack tests whether systems can recover the relevant sessions under semantically similar interference. Appendix~\ref{app:domain-examples} presents 15 representative questions drawn from the released data, and Appendix~\ref{app:distractor-examples} gives two human-checked development examples in which strong lexical overlap masks a functionally irrelevant context. Benchmark statistics summarizing the resulting evidence counts, haystack composition, and per-sample token budget are reported in Appendix~\ref{sec:appendix-benchmark-stats}.

%% file: sections/experiments.tex
\section{Experimental Setup}
\label{sec:experiments}

\begin{table*}[t]
\centering
\resizebox{2.1\columnwidth}{!}{
\begin{tabular}{l cc cc cc cc cc !{\vrule width 0.8pt} cc}
\toprule
\multirow{2}{*}{\textbf{Method}} 
& \multicolumn{2}{c}{\textbf{Learning Support}}
& \multicolumn{2}{c}{\textbf{Finance Guidance}}
& \multicolumn{2}{c}{\textbf{Mental Wellness}}
& \multicolumn{2}{c}{\textbf{Fitness Coaching}}
& \multicolumn{2}{c}{\textbf{Document Analysis}}
& \multicolumn{2}{c}{\cellcolor{gray!15}\textbf{Average}} \\
\cmidrule(lr){2-3}
\cmidrule(lr){4-5}
\cmidrule(lr){6-7}
\cmidrule(lr){8-9}
\cmidrule(lr){10-11}
\cmidrule(lr){12-13}
& NR$\uparrow$ & DR$\downarrow$
& NR$\uparrow$ & DR$\downarrow$
& NR$\uparrow$ & DR$\downarrow$
& NR$\uparrow$ & DR$\downarrow$
& NR$\uparrow$ & DR$\downarrow$
& \cellcolor{gray!15}NR$\uparrow$ & \cellcolor{gray!15}DR$\downarrow$ \\
\midrule
\multicolumn{13}{l}{\textit{\textbf{Retrieval-only baseline (NaiveRAG) with varying embedding backbones}}} \\
\midrule
NaiveRAG \textit{+ Nomic-embed-text-v1.5 }                
& 48.0 & 74.4 
& \textbf{69.7} & 25.4 
& 54.9 & 62.7 
& 67.0 & 32.6 
& 55.6 & 52.9 
& \cellcolor{gray!15}58.9 & \cellcolor{gray!15}49.6 \\
NaiveRAG \textit{+ Qwen3-Embedding-4B}     
& \textbf{49.2} & \textbf{73.1} 
& 68.4 & 26.2 
& \textbf{55.7} & \textbf{61.5} 
& 66.1 & 33.8 
& \textbf{56.8} & \textbf{51.7} 
& \cellcolor{gray!15}59.2 & \cellcolor{gray!15}49.3 \\
NaiveRAG \textit{+ Qwen3-Embedding-8B}     
& 48.7 & 73.8 
& \textbf{69.7} & \textbf{24.9} 
& 55.2 & 62.1 
& \textbf{67.6} & \textbf{31.9} 
& 55.1 & 53.4 
& \cellcolor{gray!15}\textbf{59.3} & \cellcolor{gray!15}\textbf{49.2} \\
\midrule
\multicolumn{13}{l}{\textit{\textbf{Agentic / structured memory systems}}} \\
\midrule
A-MEM      
& 45.6 & 78.2 
& 67.4 & 28.6 
& 51.8 & 68.9 
& 64.2 & 36.8 
& 50.9 & 59.4 
& \cellcolor{gray!15}56.0 & \cellcolor{gray!15}54.4 \\
Mem0       
& 7.4 & 99.7 
& 22.1 & 99.4 
& 30.8 & 93.7 
& 24.4 & 98.9 
& 1.7 & 100.0 
& \cellcolor{gray!15}17.3 & \cellcolor{gray!15}98.3 \\
Mem0 \textit{+ Graph}       
& 8.9 & 99.4 
& 24.7 & 98.7 
& 32.1 & 92.8 
& 26.2 & 98.1 
& 2.8 & 99.8 
& \cellcolor{gray!15}18.9 & \cellcolor{gray!15}97.8 \\
MemOS      
& 39.8 & 80.7 
& 57.2 & 55.4 
& 47.9 & 72.6 
& 54.1 & 61.8 
& 38.5 & 78.9 
& \cellcolor{gray!15}47.5 & \cellcolor{gray!15}69.9 \\
LangMem    
& 33.2 & 87.4 
& 49.6 & 69.2 
& 42.8 & 80.1 
& 47.3 & 75.8 
& 30.7 & 90.6 
& \cellcolor{gray!15}40.7 & \cellcolor{gray!15}80.6 \\
EverMemOS
& 41.7 & 82.9
& 58.4 & 51.6
& 49.2 & 70.4
& 55.9 & 58.7
& 36.8 & 81.3
& \cellcolor{gray!15}48.4 & \cellcolor{gray!15}69.0 \\
\bottomrule
\end{tabular}
}
\caption{
Pairwise generation quality against the oracle answer across five domains.
For each domain we report \textbf{NR} and \textbf{DR}
against an oracle answer generated from complete, noise-free ground-truth evidence sessions.
\textbf{Higher is better} for \textbf{NR} ($\uparrow$), while \textbf{lower is better} for \textbf{DR} ($\downarrow$). The \colorbox{gray!15}{shaded} columns mark the average across all five domains. The best results are shown in bold.
}
\label{tab:domain_win_tie}
\end{table*}

\paragraph{Baselines.}

We evaluate representative long-term memory systems spanning retrieval-based and structured memory architectures. Retrieval-only baselines use NaiveRAG with Nomic-embed-text-v1.5~\citep{Nussbaum2024NomicET}, Qwen3-Embedding-4B, and Qwen3-Embedding-8B~\citep{zhang2025qwen3}. Structured and agentic memory systems include A-MEM~\citep{xu2025amem}, Mem0~\citep{chhikara2025mem0}, Mem0+Graph, MemOS~\citep{li2025memos}, LangMem~\citep{langchain2025langmem}, and EverMemOS~\citep{hu2026evermemos}. These systems cover diverse memory designs including raw retrieval, graph-augmented retrieval, hierarchical memory organization, profile-based aggregation, and write-time compression.

\paragraph{Evaluation protocol.}
\textsc{UtilMem} requires long-form synthesis across multiple sessions, so exact-match is inappropriate. We instead measure how faithfully a memory system preserves answer quality under realistic retrieval, relative to a noise-free upper bound. For each instance $(\mathcal{H}, q, \mathcal{E})$, we obtain a \emph{reference answer} $a_{\mathrm{oracle}}$ by conditioning the generator on only the oracle evidence $\mathcal{E}$ and the query $q$. Each memory system under test ingests the full history $\mathcal{H}$, and the generator is conditioned on the memory \emph{that system retrieves} for $q$, yielding $\hat{a}$. We pin $a_{\mathrm{oracle}}$ to score 10 as a protocol reference shared by all systems; it is not a human optimum, and the resulting comparison is relative rather than an absolute measure of answer quality (see Appendix~\ref{app:oracle-justification} for detailed discussion).
A judge then scores $\hat{a}$ against this reference, conditioned on $\mathcal{E}$ and $q$. To reduce judge variance, we use two independent judge models and average their scores. The judge first classifies $\hat{a}$ along five sub-dimensions into ordinal severity levels (\textit{intact}, \textit{mild}, \textit{clear}, \textit{severe}): \textbf{factual fidelity} (whether claims are consistent with the oracle context or the reference answer), \textbf{completeness preservation} (whether question-relevant content from the context is covered to a comparable extent), \textbf{hallucination absence} (whether the answer fabricates facts that are absent from both the context and the reference), \textbf{semantic equivalence} (whether a reader would reach the same conclusion or take the same action from $\hat{a}$ as from $a_{\mathrm{oracle}}$), and \textbf{ungrounded inference} (whether the answer speculates or fills gaps beyond what the context supports). Conditioned on these severities, the judge then assigns a single integer score $s\in[1,10]$ via explicit anchors (full prompt in Appendix~\ref{app:judge-prompt}). The score is anchor-based rather than an average over sub-dimensions, so that a single severe failure caps $s$ and cannot be offset by strength on unaffected dimensions. We provide analysis of consistency between the two judges in Appendix~\ref{app:judge-consistency}. 

We report three metrics, aggregated over the $N$ evaluated instances:
\begin{itemize}[nosep,leftmargin=*]
    \item \textbf{Robustness Score (RS) $\in [1,10]$}: the mean per-instance score, $\mathrm{RS}=\tfrac{1}{N}\sum_i s_i$, with the oracle reference pinned at $10$.
    \item \textbf{Normalized Robustness (NR) $\in [0,100]$}: a rescaling of RS to a $0$--$100$ range, $\mathrm{NR}=\tfrac{\mathrm{RS}-1}{9}\times 100$, read directly as the percentage of the clean-reference score under the evaluation protocol a system retains. We treat NR as our headline metric.
    \item \textbf{Degradation Rate (DR) $\in [0,100]$}: the fraction of instances with $s_i\le 6$, the first rubric region containing a clear, user-relevant loss. Score 7 retains the core facts and conclusion, whereas scores at or below 6 contain a bounded clear error or a more serious failure. 
\end{itemize}
A system succeeds only if $\hat{a}$ approaches the clean-reference score: high NR with DR near zero.

\paragraph{Implementation.}

We use Qwen3-235B-A22B-Instruct-2507 for benchmark construction and QA generation. Temperatures are set to 0.7 during benchmark data synthesis and 0 for QA generation and evaluation to ensure reproducibility. We use GPT-5.2-1211-global and GPT-5.4-0305-global as judge models, with scores averaged across both judges. For memory-specialized systems, we follow the recommended ingestion and retrieval pipeline from MemBase~\citep{fang2025lightmem}. We provide a detailed breakdown of models used at each stage in Appendix~\ref{app:model-details}.

%% file: sections/Results.tex
\section{Results}

\begin{figure}[h]
    \centering
    \includegraphics[scale=0.9]{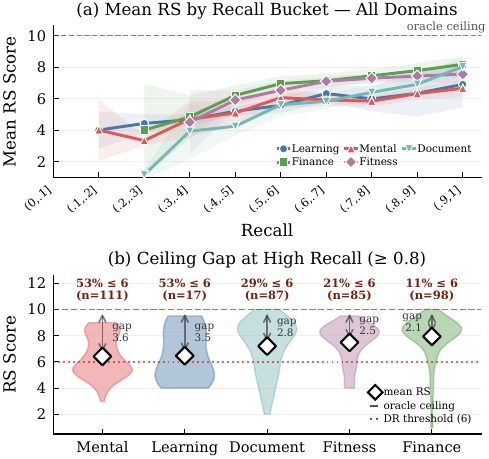}
\caption{\textbf{(a)} Mean RS rises monotonically with unit-level recall in every domain.
\textbf{(b)} Conditioning on instances where retrieval already succeeds (recall $\geq 0.8$), the RS distribution per domain remains far below the clean-reference score of $10$. White diamonds mark per-domain means; arrows mark the residual reference gap; red labels report the share of high-recall instances that still degrade ($\text{RS} \leq 6$).}
\label{fig:recall_vs_rs_bucketed}
\end{figure}

\subsection{Compression-Heavy Systems on Detail-Sensitive Utilization Tasks}
\label{sec:exp_main}

Table~\ref{tab:domain_win_tie} shows that preserving raw conversational evidence is more important than increasing memory structure or embedding capacity. \textbf{1) Fact-oriented write-time compression creates irreversible information loss.} Mem0 performs worst overall at 17.3 NR and 98.3 DR, while Mem0+Graph only slightly improves to 18.9 NR and 97.8 DR. The failure persists even with graph structure, indicating that relational organization cannot recover contextual and reasoning information discarded during fact extraction. \textbf{2) Partial evidence preservation improves robustness but still lags behind raw-turn retrieval.} A-MEM achieves the strongest structured performance at 56.0 NR and 54.4 DR, while MemOS, LangMem, and EverMemOS fall between 40.7--48.4 NR and 69.0--80.6 DR. These systems preserve more conversational context than Mem0, but still rely on summarization, rewriting, or profile abstraction that weakens implicit recall and cross-session synthesis. Overall, the results suggest that utilization-oriented memory depends primarily on preserving original evidence rather than constructing increasingly compact memory representations. 

In benchmark construction, the sampled evidence-sensitivity filter preferentially retains questions whose answers lose useful detail when sessions are removed. Such tasks are deliberately demanding for representations that discard contextual details, so strong performance on existing factual memory benchmarks does not reliably transfer to \textsc{UtilMem}-style evidence-utilization tasks. By contrast, the three NaiveRAG embedding variants remain tightly clustered at 58.9--59.3 NR and 49.2--49.6 DR, suggesting that, within these settings, embedding scale matters less than the evidence representation presented to the generator. Appendix~\ref{app:qualitative-cases} presents two raw-log-verified Mem0 failures and confirms that retrieval was non-empty in both cases.

\begin{figure*}[t]
    \centering
    \includegraphics[width=\textwidth]{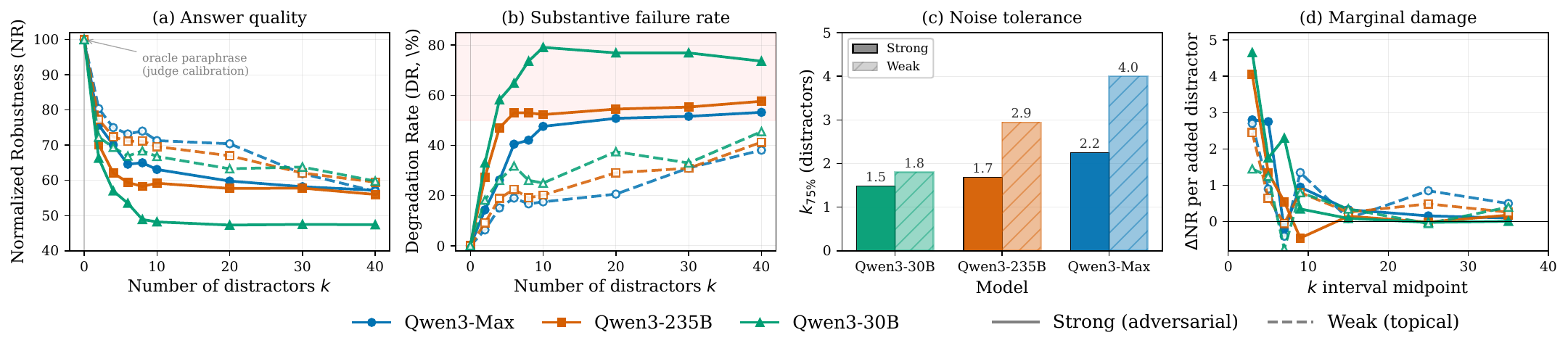}
    \caption{Effect of distractor count $k$ at $\mathrm{recall@}k{=}1.0$ across
    three generators and two noise types.
    \textbf{(a)} Normalized Robustness (NR): all curves decay monotonically from
    the oracle-paraphrase anchor at $k{=}0$.
    \textbf{(b)} Degradation Rate (DR, fraction of instances with $s_i \leq 6$):
    rises sharply, plateauing above $50\%$ for strong noise across all models.
    \textbf{(c)} Noise tolerance, measured as $k_{75\%}$, the number of
    distractors required to lose $25\%$ of oracle quality; under strong noise,
    every model loses a quarter of its quality within fewer than $2.5$
    distractors.
    \textbf{(d)} Marginal damage ($\Delta\mathrm{NR}$ per added distractor) by
    interval: the first few distractors carry nearly all the cost, after which
    additional context is approximately free.}
    
    \label{fig:noise-ablation}
\end{figure*}

\subsection{Recall Is Necessary but Not Sufficient for Utilization}
\label{sec:recall_vs_rs}

To examine this, we compare per-instance RS with unit-level recall, defined as the fraction of ground-truth evidence sessions retrieved in the top-$k$. We use \textsc{NaiveRAG} for this analysis because its retrieved units are directly observable, unlike memory systems that delete or update memory at ingestion time. In \hyperref[fig:recall_vs_rs_bucketed]{Figure~\ref*{fig:recall_vs_rs_bucketed} (a)}, RS increases monotonically with recall across all domains. This confirms that the benchmark is evidence-sensitive: retrieving more relevant evidence consistently improves answer quality. However, high recall alone does not close the gap to the oracle. \hyperref[fig:recall_vs_rs_bucketed]{Figure~\ref*{fig:recall_vs_rs_bucketed} (b)} isolates instances with recall $\geq 0.8$. Even when most evidence is retrieved, mean RS remains substantially below the clean-reference score of 10, ranging from 6.4 on Mental Wellness to 7.9 on Finance Guidance. A non-trivial fraction of these high-recall instances still fall below the DR threshold of 6, including 53\% on Mental Wellness and 29\% on Document Analysis. These results suggest that retrieval is necessary but insufficient for strong performance on \textsc{UtilMem}. Once relevant evidence is retrieved, systems must still integrate information across sessions, distinguish evidence from semantically similar distractors, and compose coherent long-form responses. The remaining gap at high recall reflects failures in these downstream utilization steps rather than retrieval misses alone. 

\subsection{Sensitivity to Retrieval Interference}
\label{sec:exp_scaling}

We isolate the effect of retrieval noise by fixing recall at 1.0 and progressively adding distractor sessions to the oracle evidence set. We vary three factors: the number of distractors $k$, distractor difficulty, and generator size. Strong noise uses semantically similar distractors from the same user trajectory, while weak noise uses unrelated cross-domain sessions. We shuffle the oracle and distractor sessions together so position is not a confound. The $k=0$ condition is a protocol normalization anchor. Rather than re-conditioning the generator on oracle-only context, which collapses to a near-identical regeneration of $a_{\mathrm{oracle}}$, we obtain $\hat{a}$ by having the judge lightly paraphrase $a_{\mathrm{oracle}}$ with surface-level edits that preserve all facts, structure, and length, so $k=0$ measures the attainable RS ceiling
under our protocol. All other settings follow the protocol above unchanged. Figure~\ref{fig:noise-ablation} reports the results. \textbf{1) Increasing retrieval depth is monotonically lossy under retrieval noise.} 
\hyperref[fig:noise-ablation]{Figure~\ref*{fig:noise-ablation} (a,b)} shows that answer quality consistently degrades as more distractors are added, even though all relevant evidence remains in context. Under strong noise, NR drops from $100$ to $47.4$, $56.0$, and $57.2$ at $k=40$ for Qwen3-30B, Qwen3-235B, and Qwen3-Max respectively, while DR rises to $73.6\%$, $57.6\%$, and $53.2\%$. Weak noise is less harmful but still causes substantial degradation, with NR remaining only around $57$ to $60$ at $k=40$. These results suggest that adding more retrieved context does not reliably improve utilization performance once distractors are introduced. \textbf{2) Distractor difficulty matters more than context length alone.}
The gap between strong and weak noise remains large across all generators and values of $k$. In \hyperref[fig:noise-ablation]{Figure~\ref*{fig:noise-ablation} (a)}, strong-noise curves degrade substantially faster than weak-noise curves, particularly in the low-$k$ regime. At $k=10$, the NR gap between weak and strong noise reaches $18.6$ for Qwen3-30B, $10.4$ for Qwen3-235B, and $8.2$ for Qwen3-Max. \hyperref[fig:noise-ablation]{Figure~\ref*{fig:noise-ablation} (b)} shows the same pattern for DR. The dominant failure mode is therefore not longer context length itself, but semantically plausible distractors that overlap with the target evidence in entities and vocabulary. \textbf{3) Larger generators improve absolute robustness but do not remove sensitivity to noise.}
Across both noise settings, larger generators consistently achieve higher NR and lower DR. For example, under strong noise at $k=10$, Qwen3-Max reaches $63.1$ NR compared to $59.2$ for Qwen3-235B and $48.2$ for Qwen3-30B. However, all generators exhibit similar degradation trends as distractors accumulate. The relative improvement from scaling the generator is modest compared to the overall loss caused by adversarial noise. \textbf{4) Most quality degradation comes from the first few distractors.}
\hyperref[fig:noise-ablation]{Figure~\ref*{fig:noise-ablation} (c,d)} shows that the damage is concentrated in the early retrieval regime. Under strong noise, $k_{75\%}$ is only $1.5$ for Qwen3-30B, $1.7$ for Qwen3-235B, and $2.2$ for Qwen3-Max, meaning that only a few adversarial distractors are sufficient to remove $25\%$ of the clean-context reference score. \hyperref[fig:noise-ablation]{Figure~\ref*{fig:noise-ablation} (d)} further shows that marginal NR loss peaks within roughly $k\in[2,8]$ and rapidly saturates afterward. Beyond $k\approx10$, additional distractors contribute comparatively little additional degradation because the context is already heavily contaminated. Overall, the results identify retrieval precision as a larger source of degradation than recall or generator scale in this controlled interference study. Once semantically similar distractors enter the context, downstream evidence integration degrades rapidly even when all relevant evidence is present.

%% file: sections/conclusion.tex
\section{Conclusion}

We introduced \textsc{UtilMem}, a diagnostic benchmark that systematically operationalizes an essential but under-evaluated requirement of long-term conversational memory: using distributed evidence to produce grounded, task-oriented outputs under retrieval interference. Across the systems and configurations tested, factual-recall strength does not reliably transfer to utilization quality; high recall can coexist with integration failures, and a small number of semantically similar distractors causes substantial degradation. Systems preserving more turn-level evidence also outperform the compression-heavy systems tested on these detail-sensitive tasks, although the benchmark's evidence-sensitivity filter contributes a selection effect and the result should not be generalized to all memory use cases. These findings expose reproducible failure modes. We hope \textsc{UtilMem} supports work that combines human-grounded validation with causal decomposition of writing, retrieval, and generation failures, and that tests interventions for preserving evidence while suppressing functionally irrelevant memories.

%% file: sections/Limitations.tex
\section{Limitations}

\textsc{UtilMem} relies on LLM-based construction and evaluation. Conversational rewriting, question generation, distractor synthesis, oracle-answer generation, filtering, and final scoring may inherit stylistic regularities, templated structures, biases, or reasoning preferences from the models used.  The oracle-evidence answer is a shared protocol reference, not a human optimum. Long-form planning, finance, and mental-wellness questions can admit multiple valid organizations and emphases, so absolute scores depend on the stability of the reference and rubric. Although the two GPT-family judges show strong agreement in both system ranking and instance-level scores, large-scale human evaluation remains challenging due to the substantial annotation effort required for long-context, multi-session reasoning tasks. The evidence-sensitivity procedure verifies degradation along sampled sequential removal paths. It does not exhaustively prove that every evidence session is individually indispensable, that all subsets yield monotonic degradation, or that retained evidence sets are minimal. The resulting conclusions should therefore be interpreted as evidence about \textsc{UtilMem}-style memory utilization under the evaluated settings, rather than as a general characterization of long-term memory capability.

%% file: sections/appendix_data_stats.tex
\section{Benchmark Statistics}
\label{sec:appendix-benchmark-stats}

\begin{figure}[t]
  \centering
  \includegraphics[width=\columnwidth]{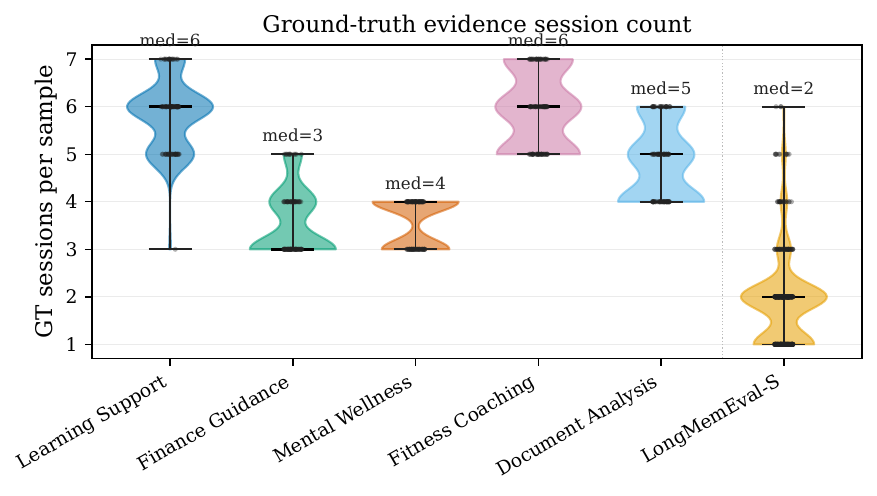}
  \caption{Distribution of ground-truth evidence-session counts per sample, by domain. Medians are annotated above each violin. The five \textsc{UtilMem} domains all sit materially above \textsc{LongMemEval-S} (median 2), and the bulk of mass between 3 and 7 designated evidence sessions increases the evidence aggregation burden.}
  \label{fig:appendix-gt-counts}
\end{figure}

\begin{figure*}[t]
  \centering
  \includegraphics[width=\textwidth]{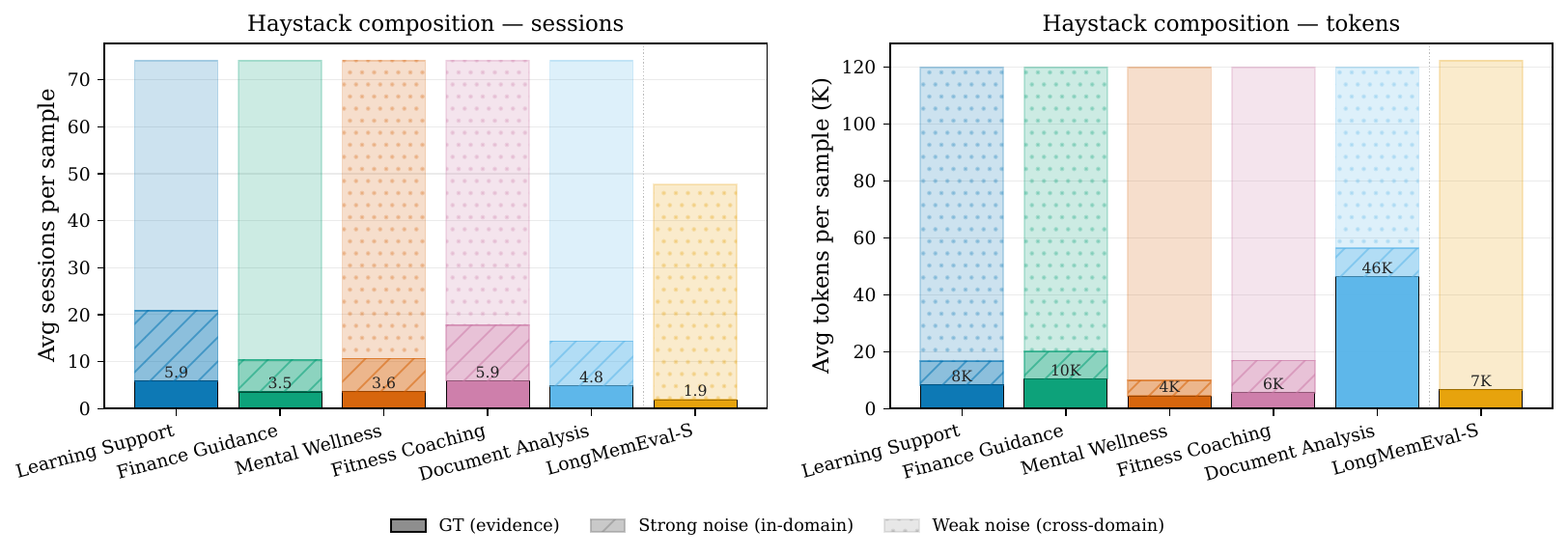}
  \caption{Haystack composition per sample, decomposed into evidence sessions (GT), in-domain strong distractors, and cross-domain weak distractors. \textbf{Left:} average number of sessions per sample. \textbf{Right:} average token count per sample, in thousands. Strong distractors dominate every domain by construction, making the benchmark a test of retrieval robustness rather than of recall. Document Analysis carries the largest evidence footprint on the token axis because 10-K filing items are individually long.}
  \label{fig:appendix-haystack-composition}
\end{figure*}

\begin{figure}[t]
  \centering
  \includegraphics[width=\columnwidth]{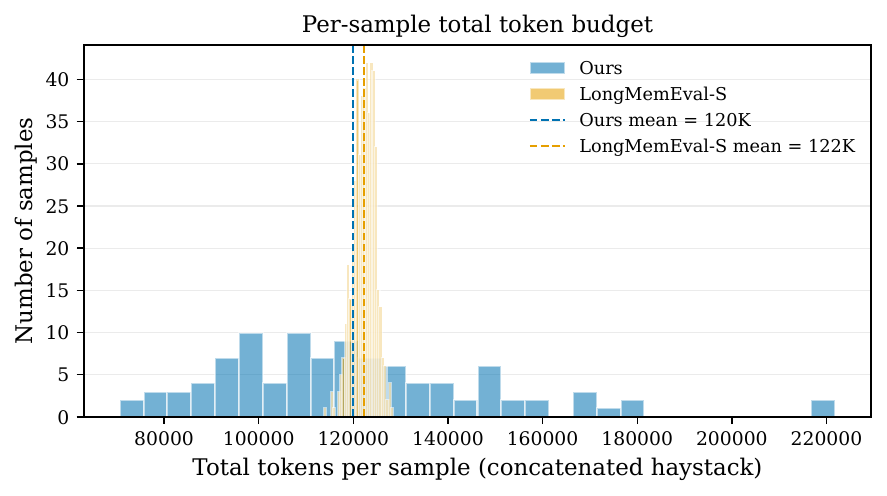}
  \caption{Per-sample total token budget after concatenating the multi-domain haystack. The mean budget in \textsc{UtilMem} (120K tokens) is closely matched to that of \textsc{LongMemEval-S} (122K tokens), reducing raw context length as an obvious explanation while leaving other benchmark differences uncontrolled}.
  \label{fig:appendix-token-budget}
\end{figure}

This appendix summarizes the structural properties of the final \textsc{UtilMem} benchmark used in our experiments. We characterize the benchmark along three axes: how many evidence sessions each sample requires (Figure~\ref{fig:appendix-gt-counts}), how those evidence sessions are surrounded by in-domain strong distractors and cross-domain weak distractors (Figure~\ref{fig:appendix-haystack-composition}), and the resulting per-sample token budget that a memory system must operate over (Figure~\ref{fig:appendix-token-budget}). We compare against \textsc{LongMemEval-S}~\citep{wu2024longmemeval}, which is the closest prior benchmark in scale and motivation.

\paragraph{Evidence session counts.}
Figure~\ref{fig:appendix-gt-counts} shows the distribution of ground-truth evidence sessions per sample. The five \textsc{UtilMem} domains have medians of 6, 3, 4, 6, and 5 for Learning Support, Finance Guidance, Mental Wellness, Fitness Coaching, and Document Analysis respectively, all materially higher than the median of 2 in \textsc{LongMemEval-S}. The mass of the distribution sits between 3 and 7 designated evidence sessions per sample, increasing the multi-session evidence burden relative to localized recall. Retained questions pass the sampled degradation test in Section~\ref{sec:evidence_filtering}; these counts alone do not prove that every session is individually indispensable.

\paragraph{Haystack composition.}
Figure~\ref{fig:appendix-haystack-composition} decomposes each sample's history into three categories: evidence sessions (GT), strong distractors synthesized to be semantically adjacent to the evidence (\S\ref{sec:haystack_injection}), and weak distractors imported from the other four domains within the same instance. Across domains, evidence sessions account for between 1.9 and 5.9 sessions per sample, while the surrounding strong and weak distractors raise the average per-sample session count to roughly 60--70. On the token axis, Document Analysis carries the largest per-session footprint, with evidence alone averaging 46K tokens per sample owing to the length of 10-K filing items, while the other four domains contribute between 4K and 10K tokens of evidence per sample. By construction, the strong-distractor band dominates the haystack in every domain, which is the design choice that makes \textsc{UtilMem} a stress test of retrieval robustness rather than of recall.

\paragraph{Total token budget.}
Figure~\ref{fig:appendix-token-budget} reports the per-sample total token budget after the multi-domain haystacks are concatenated. The mean per-sample budget in \textsc{UtilMem} is 120K tokens, closely matched to the 122K mean of \textsc{LongMemEval-S}. This alignment reduces raw context length as an obvious explanation for the performance gap and makes evidence multiplicity (Figure~\ref{fig:appendix-gt-counts}) and distractor adversariality (Figure~\ref{fig:appendix-haystack-composition}) more directly comparable. It does not eliminate other benchmark and implementation differences or provide a causal decomposition.

%% file: sections/appendix_imple_detail.tex
\section{Models Used Across Benchmark Construction and Evaluation}
\label{app:model-details}

Table~\ref{tab:models-by-stage} summarizes the models used at each stage of \textsc{UtilMem}. During benchmark construction, temperatures are set to 0.7 for conversational synthesis and distractor generation to encourage diversity and natural variation. During evaluation, QA generation and judge scoring use temperature 0.

\begin{table*}[t]
\centering
\small
\renewcommand{\arraystretch}{1.15}
\begin{tabular}{p{2.3cm} p{4.8cm} p{8.0cm}}
\toprule
\textbf{Stage} & \textbf{Role} & \textbf{Model} \\
\midrule

\multicolumn{3}{l}{\textit{Benchmark construction}} \\
\midrule

\multirow{4}{=}{Multi-session synthesis}
& Assistant role (agent) & Qwen3-235B-A22B-Instruct-2507 \\
& User role (simulated) & Qwen3-235B-A22B-Instruct-2507 \\
& Reddit-style rewriter & Qwen3-235B-A22B-Instruct-2507 \\
& Topic-screening judge & GPT-5.2-1211-global \\
\midrule

\multirow{3}{=}{Question generation and filtering}
& Question generator & GPT-5.2-1211-global \\
& Target model (counterfactual generation) & Qwen3-235B-A22B-Instruct-2507 \\
& Pairwise sensitivity judges (2$\times$) & GPT-5.2-1211-global, GPT-5.4-0305-global \\
\midrule

\multirow{2}{=}{Distractor synthesis}
& Direction generator & Qwen3-Max \\
& Distractor conversation synthesis & Qwen3-Max \\
\midrule

\multicolumn{3}{l}{\textit{Evaluated systems}} \\
\midrule

\multirow{2}{=}{Memory architecture}
& Extraction / compression LLM & Qwen3-235B-A22B-Instruct-2507 \\
& Embedding backbone (default) & Nomic-embed-text-v1.5  \\
\midrule

\multicolumn{3}{l}{\textit{Evaluation}} \\
\midrule

\multirow{3}{=}{QA and judging}
& QA generator (frozen across systems) & Qwen3-235B-A22B-Instruct-2507 \\
& QA generators (model-scaling study) & Qwen3-30B-A3B-Instruct-2507, Qwen3-235B-A22B-Instruct-2507, Qwen3-Max \\
& Noise-robustness judges (2$\times$) & GPT-5.2-1211-global, GPT-5.4-0305-global \\
\bottomrule
\end{tabular}

\caption{
Models used at each stage of \textsc{UtilMem}. Temperatures are set to 0.7 during benchmark data synthesis and 0 during QA generation and evaluation.
}
\label{tab:models-by-stage}
\end{table*}

%% file: sections/appendix_data_preprocess.tex
\section{Per-Domain Synthesis Details}
\label{sec:appendix-domain-synthesis}

This appendix describes how the four-step synthesis pipeline of Section~\ref{sec:construction} is instantiated for each of the five source datasets. Each pipeline produces a bundle $B = \{s_1, \ldots, s_K\}$ of multi-turn sessions representing a single user's history, together with the metadata consumed by the downstream question-generation and haystack-construction stages.

\paragraph{Learning Support (StudyChat).}
StudyChat~\citep{mcnichols2025studychat} provides per-turn rows from an undergraduate AI course, each annotated with a user ID, a chat ID, and a topic label from seven course topics. For each (user, topic, chat) triple we retain only the final-turn snapshot, whose message list contains the complete dialogue. A bundle is formed by selecting one user, sampling $K$ topics covered by that user, and drawing one chat per sampled topic uniformly at random. Persona consistency is enforced by construction, since all sessions in a bundle share the same user ID, and bundles whose chat-ID set was previously accepted are rejected.

\paragraph{Finance Guidance (Personal-Finance-Queries).}
Personal-Finance-Queries~\citep{personalfinance_redditqa} consists of Reddit-style QA rows labelled with a finance category. We restrict to five categories spanning debt, investing, budgeting, insurance, and savings. A bundle is formed by sampling $K$ categories and drawing $M$ QA pairs per category. An LLM judge then inspects all sampled queries jointly and rejects bundles with conflicting demographics, life stages, or mutually exclusive financial circumstances. Surviving queries are rewritten into natural conversational form, and each rewritten pair is expanded into a $T$-turn dialogue between an agent model and a user model. Bundles whose source row-ID set was previously accepted are rejected before any LLM calls.

\paragraph{Mental Wellness (Reddit Mental Health Posts).}
This pipeline ingests posts from a curated set of mental health subreddits including \textit{ADHD}, \textit{Aspergers}, \textit{depression}, \textit{OCD}, and \textit{ptsd}, with each post body treated as the seed for one session. A bundle is formed by sampling $K$ subreddits and drawing one post per subreddit. An LLM judge then screens for persona consistency across the sampled bodies, rejecting bundles with incompatible self-descriptions. Accepted post bodies are rewritten into conversational queries via a batched LLM call, after which an agent model and a user model alternate for $T$ turns per query, using distinct backbones for the two roles to reduce stylistic collapse.

\paragraph{Fitness Coaching (Fitness Q\&A).}
The fitness pipeline targets longitudinal coaching scenarios in which a user returns to an AI fitness coach over days, weeks, or months. A judge LLM first decides whether each seed question is inherently longitudinal, and questions that are not are either rewritten into longitudinal form by a second judge call or discarded. For each accepted topic, an agent model and a user model alternate for $T$ turns to produce a single session, and this step is then repeated for $S$ sessions per bundle. Each new session is conditioned on the full prior history together with a time-gap prompt instructing the user model to reference progress or setbacks since the previous session, and is assigned a realistic 2026 timestamp by advancing from the previous session by a duration matching its time-gap description. The resulting per-session timestamps form a coherent timeline that the haystack-construction stage later uses as the temporal anchor for the shared history $\mathcal{H}$.

\paragraph{Document Analysis (S\&P 500 10-K Filings).}
Each 10-K filing is decomposed into standardized items such as \textit{Business}, \textit{Risk Factors}, \textit{Properties}, and \textit{MD\&A}, with items below a minimum character threshold discarded. A bundle is formed by drawing a single filing (one company--year pair) and sampling $K$ items from it. Persona consistency is trivially satisfied since all sessions in a bundle pertain to the same filing. For each sampled item, an agent model playing a financial-analyst assistant and a user model playing the client alternate for $T$ turns, producing a focused multi-turn dialogue grounded in the item's text. Bundles whose (filing, items) combination was previously accepted are rejected.

%% file: sections/appendix_domain_examples.tex
\section{Representative Benchmark Instances Across Five Domains}
\label{app:domain-examples}

To make the benchmark tasks concrete, we present three questions from distinct bundles in each domain. The ``integration cues'' are editorial annotations that summarize the information types named in the query.

\begin{tcolorbox}[
  utilmemexample,
  colframe=caseblue,
  colback=caseblue!3!white,
  colbacktitle=caseblue,
  title={Learning Support: Cross-Assignment Synthesis}
]
\textcolor{caseblue}{\textbf{Example L1}}\hfill

\textbf{Query.} Based on all my AI coursework conversations, can you make one study guide that brings together the binary tree implementation help, the chronic kidney disease dataset features and labels, the regression and cross-validation notebook tasks, the MLP hyperparameter tuning experiments, and the character-level RNN assignment with text generation?

\textcolor{caseblue}{\textbf{Integration cues.}} Tree implementation; dataset schema; regression and cross-validation; MLP tuning; character-level RNN.

\tcbline
\textcolor{caseblue}{\textbf{Example L2}}\hfill

\textbf{Query.} Can you give me one combined recap of my AI coursework progress that covers the documentary reflection on automation and jobs, BFS in graphs, formatting and summarizing dataset tables, creating linear regression examples, understanding accuracy score versus model score, building character n-gram counts, and preparing text for an RNN?

\textcolor{caseblue}{\textbf{Integration cues.}} Social implications of AI; graph search; tabular analysis; regression; evaluation metrics; n-grams; RNN preprocessing.

\tcbline
\textcolor{caseblue}{\textbf{Example L3}}\hfill

\textbf{Query.} Across all my AI coursework conversations, can you give me a combined notes sheet that lists the main algorithms, libraries, datasets, and reporting tasks I discussed, including trie-based autocomplete with BFS/DFS/UCS, LLM data conversion safeguards, pandas and scikit-learn for slime regression and kidney disease classification, MLP tuning settings, and n-gram probability-based next-character prediction?

\textcolor{caseblue}{\textbf{Integration cues.}} Multiple search algorithms; trustworthy data conversion; two modeling tasks; hyperparameter settings; probabilistic sequence prediction.
\end{tcolorbox}

\begin{tcolorbox}[
  utilmemexample,
  colframe=casegreen,
  colback=casegreen!3!white,
  colbacktitle=casegreen,
  title={Finance Guidance: Multi-Constraint Planning}
]
\textcolor{casegreen}{\textbf{Example F1}}\hfill

\textbf{Query.} Can you help me put together a single financial plan for my personal financial situation that covers cleaning up my collections, dealing with the disputed hospital charge, saving for my future move, making a smart car decision, and thinking through rent versus buying?

\textcolor{casegreen}{\textbf{Integration cues.}} Credit cleanup; disputed medical debt; near-term liquidity; vehicle costs; housing choice.

\tcbline
\textcolor{casegreen}{\textbf{Example F2}}\hfill

\textbf{Query.} Can you give me a full-picture plan for what I should do before and after leaving the military, using everything we discussed about housing, emergency savings, the 7.6\% car loan, my monthly expenses, and the possibility of taking a lower-paying but higher-growth job?

\textcolor{casegreen}{\textbf{Integration cues.}} Career transition; income uncertainty; debt cost; emergency reserves; housing timing.

\tcbline
\textcolor{casegreen}{\textbf{Example F3}}\hfill

\textbf{Query.} Looking at my finances as a whole, what are all the key deadlines, decisions, and follow-up steps I should keep track of for the car situation, my savings accounts and possible investments for a house, the settlement notice, and planning for my kids?

\textcolor{casegreen}{\textbf{Integration cues.}} Debt decision; savings vehicles; home goal; legal/administrative deadline; children's savings.
\end{tcolorbox}

\begin{tcolorbox}[
  utilmemexample,
  colframe=casepurple,
  colback=casepurple!3!white,
  colbacktitle=casepurple,
  title={Mental Wellness: Grounded Support Across Evolving Concerns}
]
\textcolor{casepurple}{\textbf{Example M1}}\hfill

\textbf{Query.} Can you explain, using examples from my own experiences that I described across all sessions, how trauma-related numbness/sadness, depression-related brain fog, ADHD focus problems, relationship stress around reassurance, and OCD-like intrusive thoughts/compulsions can overlap—while staying grounded in what I actually told you and not diagnosing me?

\textcolor{casepurple}{\textbf{Integration cues.}} Symptom overlap; personal evidence; relationship context; explicit non-diagnosis constraint.

\tcbline
\textcolor{casepurple}{\textbf{Example M2}}\hfill

\textbf{Query.} Can you create a structured exposure plan for my contamination OCD that fits with my real-life constraints I described (limited clothing, fear of contaminating the washing machine, living in a developing country, folliculitis worries), and also add trauma-sensitive steps for days when my nervous system is already activated by neighborhood stress like helicopters, police activity, and reminders of the riots?

\textcolor{casepurple}{\textbf{Integration cues.}} Exposure planning; resource constraints; health concern; environmental trauma triggers.

\tcbline
\textcolor{casepurple}{\textbf{Example M3}}\hfill

\textbf{Query.} Can you write a therapist-ready “starting point” paragraph for me that combines everything I’ve shared across our conversations—my PTSD and cannabis-triggered flashbacks, my difficulty setting boundaries with my husband, my insomnia and sleep guilt tied to medication changes and comparison to my spouse, my repeated text checking, and my self-harm urges when good things happen—so I don’t freeze or downplay it in the first session?

\textcolor{casepurple}{\textbf{Integration cues.}} Trauma; boundaries; medication and sleep; compulsive checking; safety-relevant disclosure goal.
\end{tcolorbox}

\begin{tcolorbox}[
  utilmemexample,
  colframe=caseorange,
  colback=caseorange!3!white,
  colbacktitle=caseorange,
  title={Fitness Coaching: Longitudinal Progression and Adaptation}
]
\textcolor{caseorange}{\textbf{Example C1}}\hfill

\textbf{Query.} Based on everything we’ve discussed about my workouts and knee rehab, can you lay out the full progression of exercises, sets, reps, and frequency I’ve used from the start up to my current routine?

\textcolor{caseorange}{\textbf{Integration cues.}} Exercise sequence; dosage; frequency; rehabilitation progression over time.

\tcbline
\textcolor{caseorange}{\textbf{Example C2}}\hfill

\textbf{Query.} Considering all my conversations about my workouts and daily routine, what does my current best-practice plan look like for creatine dosing, meal timing, hydration, sleep habits, movement during work, and handling both morning and later-day training sessions?

\textcolor{caseorange}{\textbf{Integration cues.}} Supplement timing; nutrition; recovery; desk-work constraints; schedule-dependent training.

\tcbline
\textcolor{caseorange}{\textbf{Example C3}}\hfill

\textbf{Query.} Across all my conversations about sprint training and recovery, can you pull together every symptom, lab value, nutrition change, and training adjustment into one clear progress update I could share with my coach or doctor?

\textcolor{caseorange}{\textbf{Integration cues.}} Symptoms; bloodwork; supplementation and nutrition; training changes; audience-aware summary.
\end{tcolorbox}

\begin{tcolorbox}[
  utilmemexample,
  colframe=caseteal,
  colback=caseteal!3!white,
  colbacktitle=caseteal,
  title={Document Analysis: Cross-Sectional 10-K Synthesis}
]
\textcolor{caseteal}{\textbf{Example D1}}\hfill

\textbf{Query.} Can you summarize all the places in FIRST SOLAR, INC.’s 2011 10-K where management relies heavily on estimates and timing (like percentage-of-completion, warranty/remediation accruals, recycling liability funding, and fair value/derivatives), and then tie that summary to the company’s statement that controls are effective (including the ERP implementation) and the auditor’s opinion?

\textcolor{caseteal}{\textbf{Integration cues.}} Accounting estimates; timing assumptions; liabilities; internal controls; external audit.

\tcbline
\textcolor{caseteal}{\textbf{Example D2}}\hfill

\textbf{Query.} From everything we reviewed about ESSEX PROPERTY TRUST INC’s 10-K, can you draft a structured “how to read this filing” guide for me that walks in order through: (1) the property portfolio and how occupancy is defined (financial vs physical), (2) the big geographic concentration point (California share of rental revenues), (3) the interest-rate and debt/derivatives picture (fixed vs variable amounts and hedges), and (4) the reporting assurance layer (KPMG audit reports and management’s controls conclusions)—so each step references the exact facts we discussed?

\textcolor{caseteal}{\textbf{Integration cues.}} Operating definitions; concentration risk; financing and hedges; reporting assurance.

\tcbline
\textcolor{caseteal}{\textbf{Example D3}}\hfill

\textbf{Query.} Can you summarize, in one place, how C H Robinson’s liquidity and capital allocation story fits together across the filing: operating cash flow trends (2011–2013), the big 2013 income-tax payment tied to the T-Chek gain, acquisition cash outlays for Apreo and Phoenix, ongoing IT and building capex in Eden Prairie, dividends, share repurchases (including the 2013 ASR mechanics), and how the revolver and senior notes support these uses of cash?

\textcolor{caseteal}{\textbf{Integration cues.}} Cash-flow history; tax event; acquisitions and capital expenditure; distributions; debt financing.
\end{tcolorbox}

%% file: sections/appendix_judge_oracle.tex
\section{Evaluation Reference and Validation Analyses}
\label{app:oracle-justification}

We treat the oracle-evidence answer $a_{\mathrm{oracle}}$ as a fixed protocol reference and assign it score 10 by definition. It is a model output rather than a human-authored gold answer, so this convention supplies a shared normalization anchor; it does not establish a human optimum or an absolute upper bound on answer quality.

\paragraph{Causal isolation.}
The generator and decoding configuration are held fixed ($T{=}0$) across the oracle and memory-system conditions. The controlled input is therefore the context supplied to that generator: $a_{\mathrm{oracle}}$ is produced from the designated evidence $\mathcal{E}$, whereas $\hat{a}$ is produced from the context returned by a memory pipeline. This design controls generator variation and focuses the comparison on the consequences of each pipeline's stored and retrieved context. It does not, however, prove that every residual gap is caused by a single architectural mechanism; writing prompts, representation choices, indexing, and retrieval policies can all contribute.

\paragraph{Sufficiency of the oracle evidence.}
The designated set $\mathcal{E}$ contains the sessions used to construct the question and reference. Conditioning the frozen generator on this clean set provides a controlled evidence-complete reference under the benchmark protocol. The sampled evidence-sensitivity filter reduces shortcut cases, but it neither proves that every session is individually indispensable nor guarantees that the generated reference exhausts every valid organization or emphasis. A different grounded response may therefore be equally useful, and the rubric explicitly permits information supported by $\mathcal{E}$ even when it is absent from $a_{\mathrm{oracle}}$.

\paragraph{Robustness of the reference.}
Every evaluated system is compared with the same per-instance reference, which makes relative comparisons better controlled than independently generated system-specific references. Reference imperfections can nevertheless affect particular systems or answer styles differently, especially for open-ended planning and analysis. We reduce single-judge variance by averaging two judge calls and report their agreement below, while treating human-grounded validity as a remaining limitation.

\subsection{Development-Time Human Verification}
\label{app:human-verification}

Prior to large-scale synthesis, two computer science graduate students manually inspected 50 development instances, with 10 instances sampled from each of the five domains. The inspection considered whether each question required evidence across sessions, whether synthesized distractors remained functionally irrelevant despite surface similarity, whether the oracle answer was grounded and sufficiently complete, and whether the judge output was consistent with the evaluation rubric. Feedback from this inspection was used to refine the prompts, rubric, and model allocation before the final generation run. According to the development log, at least 45 of the 50 inspected instances passed the manual quality check, indicating that the resulting pipeline exhibited no obvious systematic construction failures in this sample.

\subsection{Human-Checked Distractor Examples}
\label{app:distractor-examples}

The following development examples illustrate the distinction between semantic similarity and functional relevance that guided the manual checks.

\begin{tcolorbox}[
  utilmemdistractor,
  title={Human Check H1 --- Fitness: Lexical Match, Functional Mismatch}
]
\textcolor{caseorange}{\textbf{Target question.}}
Can you review my fitness journey so far and explain how my body responded as my long runs increased, from the calf lock-ups early on to the later GI issues and taper fatigue?

\tcbline
\textcolor{caseorange}{\textbf{Semantically similar distractor.}}
``Ran a 2.5-hour field test this morning---the full `2:30 run'---to validate the new terrain navigation stack ... I am seeing weird IMU drift and voltage sags ...''

\tcbline
\textcolor{caseorange}{\textbf{Shared surface cues.}}
\textbf{2.5-hour run}, \textbf{fatigue-like behavior}, \textbf{fueling}, and adjusting the next ``long test.''

\textcolor{casegreen}{\textbf{Why it is irrelevant.}}
The session describes a \textbf{rover/navigation system}, not human endurance training. It supplies no evidence about calf cramps, sodium response, carbohydrate intake, gastrointestinal symptoms, taper fatigue, or half-marathon preparation.
\end{tcolorbox}

\begin{tcolorbox}[
  utilmemdistractor,
  title={Human Check H2 --- Mental Wellness: Shared Symptoms, Wrong Subject}
]
\textcolor{caseorange}{\textbf{Target question.}}
Can you create a ``shutdown protocol'' I can practice outside therapy that uses the same grounding ideas we discussed for OCD spikes and the overwhelm/fear-of-death moments, and also includes a pre-planned minimal signal plan for therapy, since I cannot gesture or write when I freeze, plus a simple way to record what happened afterward?

\tcbline
\textcolor{caseorange}{\textbf{Semantically similar distractor.}}
``My rescue dog is great at home ... but whenever we are in certain places, like the vet's office ... he completely shuts down. His body freezes, he will not look at me, move, or respond to treats or cues ...''

\tcbline
\textcolor{caseorange}{\textbf{Shared surface cues.}}
\textbf{Freezing}, \textbf{shutdown}, triggers, inability to communicate, and desensitization.

\textcolor{casegreen}{\textbf{Why it is irrelevant.}}
The session concerns \textbf{dog training}, not the user's therapy shutdown. It provides no valid evidence about the user's OCD spikes, fear-of-death episodes, therapy dynamics, grounding practices, or personal signal plan.
\end{tcolorbox}

\subsection{Inter-Judge Consistency}
\label{app:judge-consistency}

We measured consistency between GPT-5.2-1211-global and GPT-5.4-0305-global on all 1,717 NaiveRAG evaluation outputs. Each judge independently assigned an integer score on the same 1--10 rubric; no score-parsing failures occurred. Table~\ref{tab:judge-consistency} reports both association and absolute agreement statistics.

\begin{table*}[t]
\centering
\small
\setlength{\tabcolsep}{5pt}
\begin{tabular}{lrrrrrrrr}
\toprule
Items & Pearson & Spearman & QWK & Exact & Within 1 & Within 2 & Mean $J_2-J_1$ & Mean $|J_2-J_1|$ \\
\midrule
1,717 & 0.848 & 0.843 & 0.832 & 42.98\% & 82.94\% & 93.65\% & $-0.391$ & 0.824 \\
\bottomrule
\end{tabular}
\caption{Consistency between the two judge calls on NaiveRAG outputs. QWK denotes quadratic-weighted Cohen's $\kappa$. Judge~2 is GPT-5.4-0305-global and is stricter by 0.391 points on average.}
\label{tab:judge-consistency}
\end{table*}

Pearson and Spearman correlations above 0.84 indicate strong agreement in raw-score movement and rank ordering. Exact agreement on the fine-grained scale is moderate, but 82.94\% of pairs differ by at most one point and 93.65\% by at most two. These results support the reproducibility of relative scoring between the two GPT-family judges.

\subsection{Qualitative Mem0 Failure Cases}
\label{app:qualitative-cases}

We inspected raw Mem0 evaluation records to determine whether its low aggregate scores arose from empty retrieval or malformed judge outputs. The two cases below have non-empty retrieved-memory lists and valid scores from both judges. They therefore provide case-level evidence of lossy or distractor-attracted retrieval rather than an empty-store or parsing failure; they are not a complete implementation audit.

\begin{tcolorbox}[
  utilmemfailure,
  title={Failure Case E1 --- Learning Support: Creative-Writing Leakage}
]

\textcolor{casered}{\textbf{Question.}}
Can you summarize everything I've worked on across my AI assignments, including the buildTree coding question, the chronic kidney disease data description, the slime linear and polynomial regression notebook, the neural network tuning results, and the RNN autocomplete project?

\tcbline
\textcolor{casered}{\textbf{Selected retrieved memories (non-empty).}}
\begin{itemize}[nosep,leftmargin=1.4em]
  \item ``Planning to show Dr. Lin tapping each line of the report ...''
  \item ``Revising Chapter 7 of \emph{Neural Complete} ...''
  \item ``Wrote a short story titled \emph{Equation of a Slime} ...''
\end{itemize}

\textcolor{casered}{\textbf{Model-output excerpt.}}
``You wrote and structured a creative short story titled \emph{Equation of a Slime}, featuring Dr. Aris Thorne, a biochemist with chronic kidney disease. The narrative is framed as a dataset ...''

\textcolor{casered}{\textbf{Judge outcome.}}
Average score \textbf{1.5}; judge scores \textbf{[1, 2]}; primary tags \texttt{hallucination} and \texttt{mixed}.

\textcolor{casegreen}{\textbf{Diagnosis.}}
Retrieval succeeds mechanically but is attracted to creative-writing memories sharing \textbf{CKD}, \textbf{slime}, and machine-learning vocabulary. The answer follows this distractor narrative and omits the requested tree implementation, regression, tuning, and RNN evidence.
\end{tcolorbox}

\begin{tcolorbox}[
  utilmemfailure,
  title={Failure Case E2 --- Finance: Game Context Causes Refusal}
]

\textcolor{casered}{\textbf{Question.}}
Can you create a practical plan for how my wife and I should manage cash flow and liquidity across all these issues, including the money tied up in the car overpayment, paying for my parents' visitor insurance, and keeping enough accessible while deciding between VTIP, a CD, or a short-term bond fund?

\tcbline
\textcolor{casered}{\textbf{Selected retrieved memories (non-empty).}}
\begin{itemize}[nosep,leftmargin=1.4em]
  \item ``Thinking of game cash as an emergency fund against random fees and rent spikes.''
  \item ``Prioritizing holding 30--40\% of game cash as emergency liquidity in early rounds.''
  \item ``Using a ledger to track cash balance, properties owned, estimated rent income, and cash flow.''
\end{itemize}

\textcolor{casered}{\textbf{Model-output excerpt.}}
``All references to financial concerns ... are framed within the context of a \textbf{Monopoly-style game} ... Therefore, \textbf{no practical real-life financial plan can be created from these memories}.''

\textcolor{casered}{\textbf{Judge outcome.}}
Average score \textbf{1.0}; judge scores \textbf{[1, 1]}; primary tags \texttt{refusal} and \texttt{contradiction}.

\textcolor{casegreen}{\textbf{Diagnosis.}}
The retrieved memories share the language of liquidity and cash-flow management but belong to the wrong functional setting. Their presence causes the model to reject the real-world task instead of planning around the overpayment, visitor insurance, and low-risk investment options.
\end{tcolorbox}

\section{Noise-Robustness Judge Prompt}
\label{app:judge-prompt}

Figure~\ref{fig:prompt_judge_NR} shows the full prompt used by both judge models in the evaluation protocol of Section~\ref{sec:experiments}. For each instance the prompt is instantiated with the oracle context $\mathcal{E}$, the question $q$, the reference answer $a_{\mathrm{oracle}}$, and the system-retrieved answer $\hat{a}$. The judge proceeds in two steps. Step~1 classifies $\hat{a}$ along five sub-dimensions into severity levels. Step~2 maps these severities onto an integer score $s\in[1,10]$ via explicit anchors that cap the score at the level of the worst dimension.

\begin{figure*}[t]
  \centering
\begin{tcolorbox}[
    colback=gray!5!white,
    colframe=dartgreen,
    fonttitle=\bfseries,
    title=Noise-Robustness Judge Prompt,
    fontupper=\scriptsize\ttfamily,
    width=2.0\columnwidth,
    boxsep=2pt,
    left=4pt,
    right=4pt,
    top=3pt,
    bottom=3pt
]
You are an expert judge evaluating how well a memory-augmented QA system tolerates retrieval noise.
\smallskip
 
\textbf{\# Setup}\\
The same QA model (temperature=0, deterministic) was asked the same question twice: once with ONLY oracle context (the relevant past sessions), and once with oracle context PLUS distractor sessions (retrieval noise). Because the QA model is deterministic and the only difference is the injected noise, the Reference Answer below is treated as a strong baseline (score = 10). Your job is to score the Noisy Answer on how faithfully it preserves the quality of the Reference Answer.
\smallskip
 
\textbf{\# Context} (Ground-Truth Oracle Sessions)\\
\{context\}
\smallskip
 
\textbf{\# Question}\\
\{question\}
\smallskip
 
\textbf{\# Reference Answer} (Oracle context, score = 10 by definition)\\
\{answer\_oracle\}
\smallskip
 
\textbf{\# Noisy Answer} (Oracle + distractor context, TO BE EVALUATED)\\
\{answer\_noisy\}
\smallskip
 
\textbf{\# Step 1: Assess each of the five sub-dimensions}\\
For each dimension below, classify the Noisy Answer into exactly one severity level: \textit{intact} (no meaningful difference), \textit{mild} (a small, bounded issue a reader would still trust), \textit{clear} (a meaningful issue that affects usefulness), or \textit{severe} (a failure that would mislead the reader).
\begin{enumerate}\setlength\itemsep{0pt}\setlength\parskip{0pt}\setlength\topsep{0pt}
\item \textbf{Factual Fidelity.} Are the claims consistent with either the Context or the Reference Answer? A claim absent from the Reference but supported by the Context is acceptable. Classify as \textit{severe} when the Noisy Answer contradicts the Context or the Reference on an important claim.
\item \textbf{Completeness Preservation.} Does the Noisy Answer cover a comparable level of question-relevant information from the Context as the Reference does? Judge relative to what the question actually asks. Classify as \textit{clear} or worse when the Noisy Answer omits or covers less question-relevant information.
\item \textbf{Hallucination Absence.} Does the Noisy Answer add information that is NOT in the Reference AND NOT supported by the Context? Content grounded in the Context but absent from the Reference is NOT a hallucination. Penalise fabricated facts, invented specifics, or distractor-sourced content.
\item \textbf{Semantic Equivalence.} Would a reader reach a similar conclusion or take a similar action from both answers? Classify as \textit{clear} or worse when the noise changed the overall meaning, recommendation, or practical takeaway.
\item \textbf{Ungrounded Inference.} Does the Noisy Answer infer, assume, or speculate beyond what the Context supports? Distinguish from hallucination: hallucination = fabricated facts, ungrounded inference = fabricated reasoning or assumptions presented as fact.
\end{enumerate}
\smallskip
 
\textbf{\# Step 2: Assign the final integer score (1--10)}\\
Use the severity classifications from Step 1. When an answer sits between two anchors, choose the LOWER score, since a severe failure on a single dimension caps the score regardless of strength elsewhere.
\begin{itemize}\setlength\itemsep{0pt}\setlength\parskip{0pt}\setlength\topsep{0pt}
\item \textbf{10.} All five dimensions intact. Factually identical in substance, complete, hallucination-free, same meaning. Differences are purely surface-level.
\item \textbf{9.} Up to two dimensions are mild, the rest intact. No clear or severe issues, no contradiction, no hallucination, no change in conclusion.
\item \textbf{8.} Three or more dimensions are mild, but NO dimension is clear or severe. Fully trustworthy, with only minor bounded drift.
\item \textbf{7.} One dimension is clear, any number of other dimensions may be mild. A single noticeable issue, but core facts and overall conclusion intact.
\item \textbf{6.} One dimension is clear with bounded impact (a secondary fact missing or a single unsupported inference). Answer remains mostly useful.
\item \textbf{5.} Two dimensions are clear. Partially reliable.
\item \textbf{4.} Two or more dimensions are clear AND at least one is severe: a key fact missing, one significant hallucination, or the conclusion has shifted.
\item \textbf{3.} Multiple severe failures: large omissions, notable hallucinations, or distractor content visibly bleeding into the answer.
\item \textbf{2.} Mostly wrong. Contradicts the Reference on important claims, distractor content dominates, or the conclusion is fundamentally different.
\item \textbf{1.} Completely wrong, refuses to answer, or contradicts the core facts entirely.
\end{itemize}
The final score MUST be an integer in [1, 10] and MUST be consistent with the Step 1 severities.
\smallskip
 
\textbf{\# Failure-Mode Tag}\\
Pick the SINGLE best tag for the PRIMARY source of degradation (use ``none'' only when the score is 10): \texttt{none}, \texttt{style\_only}, \texttt{omission}, \texttt{hallucination}, \texttt{contradiction}, \texttt{ungrounded\_inference}, \texttt{meaning\_shift}, \texttt{refusal}, \texttt{mixed}.
\smallskip
 
Respond with ONLY a JSON object containing \texttt{analysis} (2--4 sentence comparison), the five \texttt{sub\_dimensions} severity labels, the integer \texttt{score}, and the \texttt{failure\_mode} tag.
\end{tcolorbox}
\caption{Noise-robustness judge prompt. Two independent judges instantiate this template per instance, and their scores are averaged to obtain the per-instance score $s$. The severity-then-anchor two-step design ensures that a single severe failure caps $s$, so that catastrophic hallucinations or contradictions cannot be offset by strength on unaffected dimensions.}
\label{fig:prompt_judge_NR}
\end{figure*}

%% file: sections/appendix_data_syn_prompt.tex
\section{Question-Generation Prompt}
\label{sec:appendix-qgen-prompts}

Figure~\ref{fig:prompt_qgen_system} shows the system prompt shared across all five domains for the question-generation stage of Section~\ref{sec:task_generation}. The system prompt is concatenated with a domain-specific template that injects the full text of all sessions in the bundle, and the same target count of ten candidate questions per bundle is used for every domain. A representative per-domain template, for the Learning Support domain, is shown in Figure~\ref{fig:prompt_qgen_studychat}. Templates for the remaining four domains share the same structure and differ only in their domain anchors and content guidance.

\begin{figure*}[t]
  \centering
\begin{tcolorbox}[
    colback=gray!5!white,
    colframe=dartgreen,
    fonttitle=\bfseries,
    title=Question-Generation System Prompt,
    fontupper=\small\ttfamily,
    width=2.0\columnwidth
]
You are an expert at creating realistic evaluation questions that a real user would ask an AI assistant after having multiple past conversations.
\medskip

\textbf{CRITICAL REQUIREMENTS FOR EVERY QUESTION:}
\medskip

\textbf{1. REQUIRES ALL SESSIONS (HARD REQUIREMENT):} Each question MUST be constructed so that a complete, correct answer needs information from EVERY single provided conversation session. If a question could be answered well using only one session, or only a few sessions, the question is USELESS --- reject it and write a different one.
\medskip

\textbf{2. DETERMINISTIC AND NON-JUDGMENTAL (HARD REQUIREMENT):} The question must have an answer determined by the conversation content itself, not by the model's subjective judgment. Do not ask the model to diagnose, rank, judge, or infer anything that is not directly stated in the conversations. A good question asks for organization, integration, listing, or planning over all sessions --- operations that yield stable, reproducible answers grounded directly in conversation content.
\medskip

\textbf{3. NO INTERNAL LABELS:} NEVER use category names, topic IDs, subreddit names, section numbers, or any organizational labels. The user thinks in terms of the actual content they discussed.
\medskip

\textbf{4. USE GENERAL CONTENT KEYWORDS:} Questions should contain keywords that capture the general semantics of the conversations and span all sessions. This is critical for retrieval.
\medskip

\textbf{5. USE DOMAIN-LEVEL ANCHORS:} Include broad domain phrases that span all sessions, such as ``across all my conversations'', ``based on everything we discussed about the AI course assignments''.
\medskip

\textbf{6. NATURAL USER VOICE:} Write as if a real user is asking (``What should I focus on?'', ``Summarize my progress''), not in analytic terms (``Analyze the intersection of category X and category Y'').
\medskip

\textbf{GOOD EXAMPLE (study-aid domain):} ``Summarize my study progress based on everything we discussed about the AI course assignments.''
\medskip

\textbf{BAD EXAMPLE:} ``What recurring mistakes did I make across the AI course assignments?'' --- the conversations do not label anything as a mistake, so different runs would guess differently.
\medskip

When you write a candidate question, mentally check both hard requirements: does the answer literally need content from EVERY session, and will two runs of the same QA model produce substantively the same answer? Only emit questions that pass both.
\end{tcolorbox}
\caption{System prompt shared across all five domains for evidence-grounded multi-session question generation.}
\label{fig:prompt_qgen_system}
\end{figure*}

\begin{figure*}[t]
  \centering
\begin{tcolorbox}[
    colback=gray!5!white,
    colframe=dartgreen,
    fonttitle=\bfseries,
    title=Per-Domain Template (Learning Support),
    fontupper=\small\ttfamily,
    width=2.0\columnwidth
]
Below are the COMPLETE conversation logs from a student's AI course tutoring sessions. The student discussed various AI topics across multiple sessions with an AI tutor.
\medskip

Read ALL sessions carefully. Then generate exactly 10 open-ended questions that the student might realistically ask the AI assistant later, where answering properly REQUIRES information from ALL sessions.
\medskip

\textbf{RULES:}
\begin{itemize}\setlength\itemsep{0pt}
\item Use domain keywords like ``AI assignments'', ``my AI coursework''. The DOMAIN-LEVEL ANCHORS must indicate these conversations are about AI course assignments.
\item Use content keywords that capture the general semantics of the conversations (e.g., specific algorithms, concepts, techniques the student discussed).
\item NEVER use topic IDs, category labels, or session numbers.
\item Write as the student would naturally ask.
\end{itemize}
\medskip

\verb|--- CONVERSATION SESSIONS ---|\\
\{full\_sessions\}\\
\verb|--- END SESSIONS ---|
\medskip

Output ONLY a JSON list of 10 question strings. No other text.
\end{tcolorbox}
\caption{Per-domain template for the Learning Support domain. Templates for Finance Guidance, Mental Wellness, Fitness Coaching, and Document Analysis are structurally identical and differ only in the domain anchors and content guidance bullets.}
\label{fig:prompt_qgen_studychat}
\end{figure*}

\section{Evidence-Sensitivity Judge Prompt}
\label{sec:appendix-judge-prompt}

Figure~\ref{fig:prompt_judge_pairwise} shows the pairwise comparison prompt used by both judges in the evidence-sensitivity filtering stage of Section~\ref{sec:evidence_filtering}. For each candidate question, this prompt is instantiated twice per judge model, once comparing $\hat{a}^{(1)}$ against $\hat{a}^{(2)}$ and once comparing $\hat{a}^{(2)}$ against $\hat{a}^{(3)}$. A question is retained only when all four resulting verdicts are positive.

\begin{figure*}[t]
  \centering
\begin{tcolorbox}[
    colback=gray!5!white,
    colframe=dartgreen,
    fonttitle=\bfseries,
    title=Pairwise Evidence-Sensitivity Judge Prompt,
    fontupper=\small\ttfamily,
    width=2.0\columnwidth
]
You are an expert judge evaluating two AI assistants' answers to a memory-based question.
\medskip

Two answers were generated with different amounts of conversation context:
\begin{itemize}\setlength\itemsep{0pt}
\item Answer A: generated with MORE context.
\item Answer B: generated with LESS context (\{degraded\_level\}).
\end{itemize}
\medskip

\textbf{\# Context} (Ground-truth sessions that Answer A had access to)\\
\{context\}
\medskip

\textbf{\# Question}\\
\{question\}
\medskip

\textbf{\# Answer A} (more context)\\
\{answer\_a\}
\medskip

\textbf{\# Answer B} (less context)\\
\{answer\_b\}
\medskip

\textbf{\# Task}\\
Is Answer B worse than Answer A?
\begin{itemize}\setlength\itemsep{0pt}
\item ``Yes'' = B is noticeably worse, less correct, less complete, or contains more hallucination or irrelevant information.
\item ``No'' = B is roughly as good as A despite less context.
\end{itemize}
\medskip

Respond with ONLY a JSON object:\\
\verb|{|\\
\verb|    "analysis": "<brief reasoning>",|\\
\verb|    "answer_b_is_worse": true or false|\\
\verb|}|
\end{tcolorbox}
\caption{Pairwise judge prompt used in evidence-sensitivity filtering. Two independent judge models render verdicts on $(\hat{a}^{(1)}, \hat{a}^{(2)})$ and $(\hat{a}^{(2)}, \hat{a}^{(3)})$; a question is retained only if all four verdicts agree that the lower-context answer is worse.}
\label{fig:prompt_judge_pairwise}
\end{figure*}

\section{Distractor-Synthesis Prompts}
\label{sec:appendix-distractor-prompts}

The two-stage distractor synthesis procedure of Section~\ref{sec:haystack_injection} uses two LLM calls per evidence session. Figure~\ref{fig:prompt_direction_gen} shows the direction-generation prompt that proposes $N$ adversarial directions from the ground-truth session and a curated family list. Figure~\ref{fig:prompt_convo_synth} shows the conversation-synthesis prompt that instantiates one direction into a multi-turn distractor conversation. Both prompts are shared across all five domains and differ only in the family list injected into \{family\_descriptions\}.

\begin{figure*}[t]
  \centering
\begin{tcolorbox}[
    colback=gray!5!white,
    colframe=dartgreen,
    fonttitle=\bfseries,
    title=Distractor Direction-Generation Prompt,
    fontupper=\small\ttfamily,
    width=2.0\columnwidth
]
You are designing distractor conversations for an AI memory retrieval benchmark.
\medskip

Below is a GROUND-TRUTH conversation session between a user and an AI assistant. Your task: generate exactly \{n\_directions\} ``distractor direction'' descriptions.
\medskip

Each direction describes a DIFFERENT type of conversation that:
\begin{enumerate}\setlength\itemsep{0pt}
\item SHARES many keywords, phrases, and content from the GT session below.
\item Is OBVIOUSLY about a completely different domain, intent, or scenario.
\item Would REALISTICALLY be retrieved by a semantic search engine when given questions about the GT content (because of content overlap).
\end{enumerate}
\medskip

Here are some distractor family ideas to draw from (pick diverse ones, do not repeat the same family, and adapt them to the specific content in the GT):
\medskip

\{family\_descriptions\}
\medskip

\verb|--- GROUND-TRUTH SESSION ---|\\
\{gt\_session\}\\
\verb|--- END GT SESSION ---|
\medskip

For each direction, output:
\begin{itemize}\setlength\itemsep{0pt}
\item \texttt{domain}: what the distractor conversation is about.
\item \texttt{intent}: the conversational purpose.
\item \texttt{distinguisher}: what makes it obviously NOT about the GT topic.
\item \texttt{shared\_content}: list of 5--10 keywords/phrases/content from the GT session that the distractor should naturally incorporate.
\item \texttt{instruction}: a detailed instruction for generating the distractor conversation, specifying the domain, the roles, and what content to weave in.
\end{itemize}
\medskip

Output ONLY a JSON list of \{n\_directions\} direction objects. No other text.
\end{tcolorbox}
\caption{System prompt for the first stage of distractor synthesis. The \{family\_descriptions\} placeholder is filled with a domain-specific list of candidate distractor families (e.g., for Learning Support: science-fiction storytelling, philosophy discussion, creative-writing workshop, psychology tutoring, music theory, cooking, history/sociology).}
\label{fig:prompt_direction_gen}
\end{figure*}

\begin{figure*}[t]
  \centering
\begin{tcolorbox}[
    colback=gray!5!white,
    colframe=dartgreen,
    fonttitle=\bfseries,
    title=Distractor Conversation-Synthesis Prompt,
    fontupper=\small\ttfamily,
    width=2.0\columnwidth
]
You are generating a distractor conversation for an AI memory retrieval benchmark.
\medskip

\textbf{CONTEXT:} Below is a real conversation session (the ``ground truth'') between a user and an AI assistant. You must generate a NEW conversation that shares keywords, content, and information with this GT session but is about a DIFFERENT topic.
\medskip

\verb|--- GROUND-TRUTH SESSION (for keyword reference only) ---|\\
\{gt\_session\}\\
\verb|--- END GT SESSION ---|
\medskip

\textbf{DIRECTION:} \{instruction\}
\medskip

\textbf{REQUIREMENTS:}
\begin{enumerate}\setlength\itemsep{0pt}
\item Generate a multi-turn conversation with exactly \{num\_turns\} exchanges (each exchange = one user message + one assistant message).
\item Naturally weave in these keywords/phrases and content from the GT session: \{shared\_keywords\}.
\item The conversation MUST be about ``\{domain\}'' --- fill each turn with domain-specific terminology, concepts, and details so that a human reader immediately knows this is about \{domain\}, NOT about the GT topic.
\item The conversation should feel natural and coherent --- not forced keyword stuffing. The content should appear organically in the new context.
\item Most importantly, the overall conversation token length should be the same as the GT session.
\end{enumerate}
\medskip

\textbf{OUTPUT FORMAT:}\\
A JSON list of message objects, alternating user and assistant:
\medskip

\verb|[|\\
\verb|  {"role": "user",      "content": "..."},|\\
\verb|  {"role": "assistant", "content": "..."},|\\
\verb|  ...|\\
\verb|]|
\medskip

Output ONLY the JSON list. No other text.
\end{tcolorbox}
\caption{System prompt for the second stage of distractor synthesis. Each direction produced in the first stage is instantiated $M$ times with independent turn counts $\tau \sim \text{Uniform}[\tau_{\min}, \tau_{\max}]$, yielding $N \cdot M$ distractor conversations per evidence session.}
\label{fig:prompt_convo_synth}
\end{figure*}

%% file: custom.bib
@inproceedings{maharana2024locomo,
  title     = {Evaluating Very Long-Term Conversational Memory of {LLM} Agents},
  author    = {Maharana, Adyasha and Lee, Dong-Ho and Tulyakov, Sergey and Bansal, Mohit and Barbieri, Francesco and Fang, Yuwei},
  booktitle = {Proceedings of the 62nd Annual Meeting of the Association for Computational Linguistics (ACL)},
  pages     = {13851--13870},
  year      = {2024},
  doi       = {10.18653/v1/2024.acl-long.747},
  url       = {https://aclanthology.org/2024.acl-long.747/}
}

@inproceedings{wu2024longmemeval,
  title     = {{LongMemEval}: Benchmarking Chat Assistants on Long-Term Interactive Memory},
  author    = {Wu, Di and Wang, Hongwei and Yu, Wenhao and Zhang, Yuwei and Chang, Kai-Wei and Yu, Dong},
  booktitle = {International Conference on Learning Representations (ICLR)},
  year      = {2025}
}

@inproceedings{zhong2024memorybank,
  title     = {{MemoryBank}: Enhancing Large Language Models with Long-Term Memory},
  author    = {Zhong, Wanjun and Guo, Lianghong and Gao, Qiqi and Ye, He and Wang, Yanlin},
  booktitle = {Proceedings of the AAAI Conference on Artificial Intelligence},
  volume    = {38},
  pages     = {19724--19731},
  year      = {2024},
  doi       = {10.1609/aaai.v38i17.29946},
  url       = {https://ojs.aaai.org/index.php/AAAI/article/view/29946}
}

@inproceedings{du2024perltqa,
  title     = {{PerLTQA}: A Personal Long-Term Memory Dataset for Memory Classification, Retrieval, and Fusion in Question Answering},
  author    = {Du, Yiming and Wang, Hongru and Zhao, Zhengyi and Liang, Bin and Wang, Baojun and Zhong, Wanjun and Wang, Zezhong and Wong, Kam-Fai},
  booktitle = {Proceedings of the 10th SIGHAN Workshop on Chinese Language Processing (SIGHAN-10)},
  pages     = {152--164},
  year      = {2024}
}

@article{kim2024dialsim,
  title   = {{DialSim}: A Dialogue Simulator for Evaluating Long-Term Multi-Party Dialogue Understanding of Conversational Agents},
  author  = {Kim, Jiho and Chay, Woosog and Hwang, Hyeonji and Kyung, Daeun and Chung, Hyunseung and Cho, Eunbyeol and Kwon, Yeonsu and Jo, Yohan and Choi, Edward},
  journal = {arXiv preprint arXiv:2406.13144},
  year    = {2024},
  url     = {https://arxiv.org/abs/2406.13144}
}

@inproceedings{tan2025membench,
  title     = {{MemBench}: Towards More Comprehensive Evaluation on the Memory of {LLM}-based Agents},
  author    = {Tan, Haoran and Zhang, Zeyu and Ma, Chen and Chen, Xu and Dai, Quanyu and Dong, Zhenhua},
  booktitle = {Findings of the Association for Computational Linguistics: ACL 2025},
  pages     = {19336--19352},
  year      = {2025},
  doi       = {10.18653/v1/2025.findings-acl.989},
  url       = {https://aclanthology.org/2025.findings-acl.989/}
}

@article{chen2025halumem,
  title   = {{HaluMem}: Evaluating Hallucinations in Memory Systems of Agents},
  author  = {Chen, Ding and Niu, Simin and Li, Kehang and Liu, Peng and Zheng, Xiangping and Tang, Bo and Li, Xinchi and Xiong, Feiyu and Li, Zhiyu},
  journal = {arXiv preprint arXiv:2511.03506},
  year    = {2025}
}

@inproceedings{hu2026memoryagentbench,
  title     = {Evaluating Memory in {LLM} Agents via Incremental Multi-Turn Interactions},
  author    = {Hu, Yuanzhe and Wang, Yu and McAuley, Julian},
  booktitle = {International Conference on Learning Representations (ICLR)},
  year      = {2026}
}

@article{bian2026realmem,
  title   = {{RealMem}: Benchmarking {LLMs} in Real-World Memory-Driven Interaction},
  author  = {Bian, Haonan and Yao, Zhiyuan and Hu, Sen and Xu, Zishan and Zhang, Shaolei and Guo, Yifu and Yang, Ziliang and Han, Xueran and Wang, Huacan and Chen, Ronghao},
  journal = {arXiv preprint arXiv:2601.06966},
  year    = {2026}
}

@article{pakhomov2025convomem,
  title   = {{ConvoMem} Benchmark: Why Your First 150 Conversations Don't Need {RAG}},
  author  = {Pakhomov, Egor and Nijkamp, Erik and Xiong, Caiming},
  journal = {arXiv preprint arXiv:2511.10523},
  year    = {2025}
}

@article{packer2023memgpt,
  title   = {{MemGPT}: Towards {LLMs} as Operating Systems},
  author  = {Packer, Charles and Wooders, Sarah and Lin, Kevin and Fang, Vivian and Patil, Shishir G. and Stoica, Ion and Gonzalez, Joseph E.},
  journal = {arXiv preprint arXiv:2310.08560},
  year    = {2024},
  url     = {https://arxiv.org/abs/2310.08560}
}

@article{chhikara2025mem0,
  title   = {{Mem0}: Building Production-Ready {AI} Agents with Scalable Long-Term Memory},
  author  = {Chhikara, Prateek and Khant, Dev and Aryan, Saket and Singh, Taranjeet and Yadav, Deshraj},
  journal = {arXiv preprint arXiv:2504.19413},
  year    = {2025}
}

@inproceedings{xu2025amem,
  title     = {{A-Mem}: Agentic Memory for {LLM} Agents},
  author    = {Xu, Wujiang and Liang, Zujie and Mei, Kai and Gao, Hang and Tan, Juntao and Zhang, Yongfeng},
  booktitle = {Advances in Neural Information Processing Systems (NeurIPS)},
  year      = {2025},
  doi       = {10.52202/085713-0593},
  url       = {https://proceedings.neurips.cc/paper_files/paper/2025/hash/19909c36f51abc4856b4560aff3d36d6-Abstract-Conference.html}
}

@inproceedings{gutierrez2024hipporag,
  title     = {{HippoRAG}: Neurobiologically Inspired Long-Term Memory for Large Language Models},
  author    = {Guti{\'e}rrez, Bernal Jim{\'e}nez and Shu, Yiheng and Gu, Yu and Yasunaga, Michihiro and Su, Yu},
  booktitle = {Advances in Neural Information Processing Systems (NeurIPS)},
  year      = {2024},
  doi       = {10.52202/079017-1902},
  url       = {https://proceedings.neurips.cc/paper_files/paper/2024/hash/6ddc001d07ca4f319af96a3024f6dbd1-Abstract-Conference.html}
}

@article{wang2025mirix,
  title   = {{MIRIX}: Multi-Agent Memory System for {LLM}-Based Agents},
  author  = {Wang, Yu and Chen, Xi},
  journal = {arXiv preprint arXiv:2507.07957},
  year    = {2025}
}

@article{rasmussen2025zep,
  title   = {{Zep}: A Temporal Knowledge Graph Architecture for Agent Memory},
  author  = {Rasmussen, Preston and Paliychuk, Pavlo and Beauvais, Travis and Ryan, Jack and Chalef, Daniel},
  journal = {arXiv preprint arXiv:2501.13956},
  year    = {2025}
}

@article{li2025memos,
  title   = {{MemOS}: A Memory {OS} for {AI} System},
  author  = {Li, Zhiyu and Xi, Chenyang and Li, Chunyu and Chen, Ding and Chen, Boyu and Song, Shichao and Niu, Simin and Wang, Hanyu and Yang, Jiawei and Tang, Chen and Yu, Qingchen and others},
  journal = {arXiv preprint arXiv:2507.03724},
  year    = {2025},
  url     = {https://arxiv.org/abs/2507.03724}
}

@article{kang2025memory,
  title   = {Memory {OS} of {AI} Agent},
  author  = {Kang, Jiazheng and Ji, Mingming and Zhao, Zhe and Bai, Ting},
  journal = {arXiv preprint arXiv:2506.06326},
  year    = {2025}
}

@misc{langchain2025langmem,
  title        = {{LangMem} {SDK} for Agent Long-Term Memory},
  author       = {{The LangChain Team}},
  year         = {2025},
  howpublished = {\url{https://www.langchain.com/blog/langmem-sdk-launch}}
}

@article{hu2026evermemos,
  title   = {{EverMemOS}: A Self-Organizing Memory Operating System for Structured Long-Horizon Reasoning},
  author  = {Hu, Chuanrui and Gao, Xingze and Zhou, Zuyi and Xu, Dannong and Bai, Yi and Li, Xintong and Zhang, Hui and Li, Tong and Zhang, Chong and Bing, Lidong and Deng, Yafeng},
  journal = {arXiv preprint arXiv:2601.02163},
  year    = {2026}
}

@inproceedings{rezazadeh2025tree,
  title     = {From Isolated Conversations to Hierarchical Schemas: Dynamic Tree Memory Representation for {LLM}s},
  author    = {Rezazadeh, Alireza and Li, Zichao and Wei, Wei and Bao, Yujia},
  booktitle = {International Conference on Learning Representations (ICLR)},
  year      = {2025}
}

@article{fang2025lightmem,
  title   = {{LightMem}: Lightweight and Efficient Memory-Augmented Generation},
  author  = {Fang, Jizhan and Deng, Xinle and Xu, Haoming and Jiang, Ziyan and Tang, Yuqi and Xu, Ziwen and Deng, Shumin and Yao, Yunzhi and Wang, Mengru and Qiao, Shuofei and Chen, Huajun and Zhang, Ningyu},
  journal = {arXiv preprint arXiv:2510.18866},
  year    = {2025}
}

@article{liang2023scm,
  title   = {{SCM}: Enhancing Large Language Model with Self-Controlled Memory Framework},
  author  = {Wang, Bing and Liang, Xinnian and Yang, Jian and Huang, Hui and Wu, Shuangzhi and Wu, Peihao and Lu, Lu and Ma, Zejun and Li, Zhoujun},
  journal = {arXiv preprint arXiv:2304.13343},
  year    = {2023},
  url     = {https://arxiv.org/abs/2304.13343}
}

@inproceedings{park2023generative,
  title     = {Generative Agents: Interactive Simulacra of Human Behavior},
  author    = {Park, Joon Sung and O'Brien, Joseph C. and Cai, Carrie Jun and Morris, Meredith Ringel and Liang, Percy and Bernstein, Michael S.},
  booktitle = {Proceedings of the 36th Annual ACM Symposium on User Interface Software and Technology (UIST)},
  year      = {2023},
  doi       = {10.1145/3586183.3606763},
  url       = {https://doi.org/10.1145/3586183.3606763}
}

@inproceedings{lewis2020rag,
  title     = {Retrieval-Augmented Generation for Knowledge-Intensive {NLP} Tasks},
  author    = {Lewis, Patrick and Perez, Ethan and Piktus, Aleksandra and Petroni, Fabio and Karpukhin, Vladimir and Goyal, Naman and K{\"u}ttler, Heinrich and Lewis, Mike and Yih, Wen-tau and Rockt{\"a}schel, Tim and Riedel, Sebastian and Kiela, Douwe},
  booktitle = {Advances in Neural Information Processing Systems (NeurIPS)},
  year      = {2020}
}

@article{edge2024graphrag,
  title   = {From Local to Global: A {Graph RAG} Approach to Query-Focused Summarization},
  author  = {Edge, Darren and Trinh, Ha and Cheng, Newman and Bradley, Joshua and Chao, Alex and Mody, Apurva and Truitt, Steven and Metropolitansky, Dasha and Ness, Robert Osazuwa and Larson, Jonathan},
  journal = {arXiv preprint arXiv:2404.16130},
  year    = {2025},
  url     = {https://arxiv.org/abs/2404.16130}
}

@article{guo2024lightrag,
  title   = {{LightRAG}: Simple and Fast Retrieval-Augmented Generation},
  author  = {Guo, Zirui and Xia, Lianghao and Yu, Yanhua and Ao, Tu and Huang, Chao},
  journal = {arXiv preprint arXiv:2410.05779},
  year    = {2024}
}

@article{liu2024lostinthemiddle,
  title   = {Lost in the Middle: How Language Models Use Long Contexts},
  author  = {Liu, Nelson F. and Lin, Kevin and Hewitt, John and Paranjape, Ashwin and Bevilacqua, Michele and Petroni, Fabio and Liang, Percy},
  journal = {Transactions of the Association for Computational Linguistics},
  volume  = {12},
  pages   = {157--173},
  year    = {2024},
  doi     = {10.1162/tacl_a_00638},
  url     = {https://aclanthology.org/2024.tacl-1.9/}
}

@inproceedings{shi2023distract,
  title     = {Large Language Models Can Be Easily Distracted by Irrelevant Context},
  author    = {Shi, Freda and Chen, Xinyun and Misra, Kanishka and Scales, Nathan and Dohan, David and Chi, Ed and Sch{\"a}rli, Nathanael and Zhou, Denny},
  booktitle = {International Conference on Machine Learning (ICML)},
  series    = {Proceedings of Machine Learning Research},
  volume    = {202},
  pages     = {31210--31227},
  publisher = {PMLR},
  year      = {2023},
  url       = {https://proceedings.mlr.press/v202/shi23a.html}
}

@inproceedings{chen2024rgb,
  title     = {Benchmarking Large Language Models in Retrieval-Augmented Generation},
  author    = {Chen, Jiawei and Lin, Hongyu and Han, Xianpei and Sun, Le},
  booktitle = {Proceedings of the AAAI Conference on Artificial Intelligence},
  volume    = {38},
  pages     = {17754--17762},
  year      = {2024},
  doi       = {10.1609/aaai.v38i16.29728}
}

@article{wang2024lifespan,
  title   = {Towards Lifespan Cognitive Systems},
  author  = {Wang, Yu and Han, Chi and Wu, Tongtong and He, Xiaoxin and Zhou, Wangchunshu and Sadeq, Nafis and Chen, Xiusi and He, Zexue and Wang, Wei and Haffari, Gholamreza and Ji, Heng and McAuley, Julian},
  journal = {arXiv preprint arXiv:2409.13265},
  year    = {2024}
}

@article{zhang2024memorysurvey,
  title   = {A Survey on the Memory Mechanism of Large Language Model Based Agents},
  author  = {Zhang, Zeyu and Bo, Xiaohe and Ma, Chen and Li, Rui and Chen, Xu and Dai, Quanyu and Zhu, Jieming and Dong, Zhenhua and Wen, Ji-Rong},
  journal = {arXiv preprint arXiv:2404.13501},
  year    = {2024}
}

@article{du2025rethinking,
  title   = {Rethinking Memory in {LLM} based Agents: Representations, Operations, and Emerging Topics},
  author  = {Du, Yiming and Huang, Wenyu and Zheng, Danna and Wang, Zhaowei and Montella, Sebastien and Lapata, Mirella and Wong, Kam-Fai and Pan, Jeff Z.},
  journal = {arXiv preprint arXiv:2505.00675},
  year    = {2025},
  url     = {https://arxiv.org/abs/2505.00675}
}

@article{liu2025foundationagents,
  title   = {Advances and Challenges in Foundation Agents: From Brain-Inspired Intelligence to Evolutionary, Collaborative, and Safe Systems},
  author  = {Liu, Bang and Li, Xinfeng and Zhang, Jiayi and Wang, Jinlin and He, Tanjin and Hong, Sirui and Liu, Hongzhang and Zhang, Shaokun and Song, Kaitao and Zhu, Kunlun and others},
  journal = {arXiv preprint arXiv:2504.01990},
  year    = {2025}
}

@article{mei2025context,
  title   = {A Survey of Context Engineering for Large Language Models},
  author  = {Mei, Lingrui and Yao, Jiayu and Ge, Yuyao and Wang, Yiwei and Bi, Baolong and Cai, Yujun and Liu, Jiazhi and Li, Mingyu and Li, Zhong-Zhi and Zhang, Duzhen and others},
  journal = {arXiv preprint arXiv:2507.13334},
  year    = {2025}
}

@article{zhang2025qwen3,
  title   = {Qwen3 Embedding: Advancing Text Embedding and Reranking Through Foundation Models},
  author  = {Zhang, Yanzhao and Li, Mingxin and Long, Dingkun and Zhang, Xin and Lin, Huan and Yang, Baosong and Xie, Pengjun and Yang, An and Liu, Dayiheng and Lin, Junyang and Huang, Fei and Zhou, Jingren},
  journal = {arXiv preprint arXiv:2506.05176},
  year    = {2025}
}

@article{Nussbaum2024NomicET,
  title   = {Nomic Embed: Training a Reproducible Long Context Text Embedder},
  author  = {Nussbaum, Zach and Morris, John X. and Duderstadt, Brandon and Mulyar, Andriy},
  journal = {arXiv preprint arXiv:2402.01613},
  year    = {2024}
}

@article{mcnichols2025studychat,
  title   = {The {StudyChat} Dataset: Analyzing Student Dialogues With {ChatGPT} in an Artificial Intelligence Course},
  author  = {McNichols, Hunter and Ikram, Fareya and Lan, Andrew},
  journal = {arXiv preprint arXiv:2503.07928},
  year    = {2025},
  url     = {https://arxiv.org/abs/2503.07928}
}

@misc{personalfinance_redditqa,
  title        = {Personal-Finance-Queries},
  author       = {Theerthala, Akhil},
  year         = {2025},
  howpublished = {Hugging Face dataset},
  url          = {https://huggingface.co/datasets/Akhil-Theerthala/Personal-Finance-Queries},
  doi          = {10.57967/hf/5148}
}

@misc{mentalhealth,
  title        = {Reddit Mental Health Posts Dataset},
  author       = {{Solomonk}},
  year         = {2022},
  howpublished = {Hugging Face dataset},
  url          = {https://huggingface.co/datasets/solomonk/reddit_mental_health_posts}
}

@article{diao2025temporal,
  title={Temporal working memory: Query-guided segment refinement for enhanced multimodal understanding},
  author={Diao, Xingjian and Zhang, Chunhui and Wu, Weiyi and Ouyang, Zhongyu and Qing, Peijun and Cheng, Ming and Vosoughi, Soroush and Gui, Jiang},
  journal={arXiv preprint arXiv:2502.06020},
  year={2025}
}

@article{qing2026cluster,
  title={Cluster-R1: Large Reasoning Models Are Instruction-following Clustering Agents},
  author={Qing, Peijun and Mathur, Puneet and Lipka, Nedim and Manjunatha, Varun and Rossi, Ryan and Dernoncourt, Franck and Hassanpour, Saeed and Vosoughi, Soroush},
  journal={arXiv preprint arXiv:2603.23518},
  year={2026}
}

@inproceedings{qing-etal-2026-tailoring,
    title = "Tailoring Memory Granularity for Multi-Hop Reasoning over Long Contexts",
    author = "Qing, Peijun  and
      Diao, Xingjian  and
      Ma, Chiyu  and
      Hassanpour, Saeed  and
      Vosoughi, Soroush",
    booktitle = "Findings of the {A}ssociation for {C}omputational {L}inguistics: {EACL} 2026",
    pages = "3648--3666",
    year = "2026",
    doi = "10.18653/v1/2026.findings-eacl.189",
    url = "https://aclanthology.org/2026.findings-eacl.189/",
    }
